\pdfoutput=1 \PassOptionsToPackage{table}{xcolor}
\PassOptionsToPackage{numbers,sort&compress}{natbib}
\documentclass{article} \usepackage[preprint]{neurips_2026}
\usepackage[utf8]{inputenc}
\usepackage[T1]{fontenc}

\usepackage{amsmath,amsfonts,bm}

\def\eqref#1{equation~\ref{#1}}

\def\1{\bm{1}}

\DeclareMathAlphabet{\mathsfit}{\encodingdefault}{\sfdefault}{m}{sl}
\SetMathAlphabet{\mathsfit}{bold}{\encodingdefault}{\sfdefault}{bx}{n}

 \newcommand{\FPTaskCount}{56}
\newcommand{\FPModelCount}{26}

\newcommand{\FPAstraZeroFilled}{0}

\newcommand{\FPTerraZeroFilled}{0}

\newcommand{\FPLunaZeroFilled}{1}

\newcommand{\FPSolZeroFilled}{0}

\newcommand{\FPGrokZeroFilled}{0}

\newcommand{\FPGrokFourSevenZeroFilled}{0}

\newcommand{\FPKimiZeroFilled}{0}

\newcommand{\FPFableZeroFilled}{0}

\newcommand{\FPGeminiZeroFilled}{0}

\newcommand{\FPDeepSeekZeroFilled}{0}

\newcommand{\FPOpusFiveFiveOverall}{27.4\%}

\newcommand{\FPOpusFiveFivePasses}{46}

\newcommand{\FPOpusFiveFiveZeroFilled}{0}

\newcommand{\FPHaikuFiveFiveZeroFilled}{0}

\newcommand{\FPOpusFiveZeroFilled}{0}

\newcommand{\FPGLMZeroFilled}{0}

\newcommand{\FPSixOneSolZeroFilled}{0}

\newcommand{\FPAstraMaxZeroFilled}{0}

\newcommand{\FPSixSolZeroFilled}{0}

\newcommand{\FPSolMaxZeroFilled}{0}

\newcommand{\FPLunaMaxZeroFilled}{0}

\newcommand{\FPTerraMaxZeroFilled}{0}

\newcommand{\FPMiMoZeroFilled}{0}

\newcommand{\FPGeminiFourZeroFilled}{0}

\newcommand{\FPMistralLargeFourZeroFilled}{0}

\newcommand{\FPKimiThreeOneProZeroFilled}{0}

\newcommand{\FPDeepSeekFourOneFlashZeroFilled}{4}

\newcommand{\FPSixOneTerraZeroFilled}{0}

\usepackage{float}
\usepackage[colorlinks,breaklinks=true,plainpages=false,
  citecolor=blue,linkcolor=blue,urlcolor=blue,bookmarksdepth=3]{hyperref}
\usepackage{url}
\usepackage{xurl}\AtBeginDocument{}
\newcommand*{\doi}[1]{doi:~\href{https://doi.org/#1}{\nolinkurl{#1}}}
\usepackage{graphicx}
\usepackage{capt-of}
\usepackage{booktabs}
\usepackage{tabularx}
\usepackage{longtable}
\usepackage{array}
\usepackage{amssymb}
\usepackage[table]{xcolor}
\definecolor{altrow}{RGB}{246,246,249}
\definecolor{ours}{HTML}{EBF4EF}

\newcolumntype{Y}{>{\centering\arraybackslash}X}

\makeatletter
\def\@titlelogo{}
\newcommand\titlelogo[1]{\gdef\@titlelogo{#1}}
\def\@affiliations{}
\newcommand\affiliations[1]{\gdef\@affiliations{#1}}
\def\@titlenotes{}
\newcommand\titlenotes[1]{\gdef\@titlenotes{#1}}
\renewcommand{\@maketitle}{\vbox{\hsize\textwidth
    \linewidth\hsize
    \vskip 0.04in
    \hrule height 4\p@
    \vskip 0.12in
    \vskip -\parskip
    \centering
    {\LARGE\bf
      \ifx\@titlelogo\@empty\else
\raisebox{\dimexpr.5\fontcharht\font`H-.5\height\relax}{\includegraphics[height=19pt]{\@titlelogo}}\hspace{6pt}\fi
      \@title\par}
    \vskip 0.1in
    \vskip -\parskip
    \hrule height 1\p@
    \vskip 0.045in
    {\small\bfseries\@author\par}
    \vskip 0.04in
  }\fp@titlefootnote}
\def\fp@titlefootnote{\ifx\@affiliations\@empty\ifx\@titlenotes\@empty\else\fp@titlefootnote@\fi
  \else\fp@titlefootnote@\fi}
\def\fp@titlefootnote@{\begingroup
\interfootnotelinepenalty\@M
  \let\@thefnmark\@empty\long\def\@makefntext##1{\parindent\z@\noindent##1}\@footnotetext{\scriptsize\hyphenpenalty\@M\@affiliations\ifx\@titlenotes\@empty\else\ \@titlenotes\fi}\endgroup}
\renewcommand{\@notice}{}
\renewenvironment{abstract}{\vskip 0.05in\centerline{\large\bf Abstract}\vspace{0.4ex}\list{}{\leftmargin1.5em\rightmargin\leftmargin}\item\relax}{\par\endlist\vskip 0.5ex}
\makeatother

\title{FrontierPhysics: Evaluating Agents for \\End-to-End Physics Research}
\titlelogo{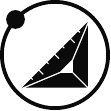}

\newcommand{\fpauthor}[2][]{\mbox{#2\ifx\relax#1\relax\else\textsuperscript{#1}\fi}}
\newcommand{\fpaffil}[2]{\textsuperscript{#1}#2}
\author{\fpauthor[1,2*]{Bingran You},
  \fpauthor[3*]{Wenjun Ke},
  \fpauthor[4*]{Ria Rosenauer},
  \fpauthor[5,6*]{Bo Liu},
  \fpauthor[7\dag\S]{Wenbo Chen},
  \fpauthor[8\dag]{Min Chen},
  \fpauthor[2\dag]{Shuqi Xu},\\
  \fpauthor[9]{Yue Ma},
  \fpauthor[2]{Jiechao Feng},
  \fpauthor[10]{Bo Chen},
  \fpauthor[3]{Xiangkai Sun},
  \fpauthor[2]{Jiarui Liu},
  \fpauthor[10]{Jinen Guo},
  \fpauthor[11,12]{Zeyu Zhou},\\
  \fpauthor[2]{Xu Huang},
  \fpauthor[13]{Zhuohan Li},
  \fpauthor[14]{Haoyu Li},
  \fpauthor[15]{Heguang Lin},
  \fpauthor[16]{Lijie Ding},
  \fpauthor[17]{Yadong Zeng},
  \fpauthor[18]{Kangqi Zhang},\\
  \fpauthor[19]{Xinge Liu},
  \fpauthor[20]{Hanwen Liu},
  \fpauthor[10]{Zhenghong Li},
  \fpauthor[6,21]{Alexander von Recum},
  \fpauthor[12]{Yuchen Zhu},\\
  \fpauthor[21]{Felix Schwarzfischer},
  \fpauthor[22]{Yizhou Zhao},
  \fpauthor[23]{Jicheng Wang},
  \fpauthor[24]{Binxu Li},
  \fpauthor[2]{Qian Yu},
  \fpauthor[9]{Chan Li},\\
  \fpauthor[10]{Cheng Dai},
  \fpauthor[10]{Di Wang},
  \fpauthor[25]{Haoyu Zhang},
  \fpauthor[6]{Jiankai Sun},
  \fpauthor[10]{Kaituo Feng},
  \fpauthor[26]{Mian Qin},
  \fpauthor[27]{Ruihong Yin},\\
  \fpauthor[22]{Yiqun Luo},
  \fpauthor[6]{Weiwei Wu},
  \fpauthor[28]{Shihong Pan},
  \fpauthor[22]{Tsung Yeh Hsieh},
  \fpauthor[29]{Xiao Huang},
  \fpauthor[10]{Yan Zhang},\\
  \fpauthor[30]{Sihan Li},
  \fpauthor[31]{Zirui Ren},
  \fpauthor[31]{Zengji Tu},
  \fpauthor[32]{Kyoung Whan Choe},
  \fpauthor[2]{Ke Sun},
  \fpauthor[33]{Vincenzo Galia},\\
  \fpauthor[34]{Vaisakhi Mishra},
  \fpauthor[26]{Jiashu He},
  \fpauthor[7\S]{Hongyang Wang},
  \fpauthor[35]{Zhiheng Yang},
  \fpauthor[36]{Shubham Gaur},\\
  \fpauthor[7\S]{Yifan Gu},
  \fpauthor[10]{Shuyi Wang},
  \fpauthor[14]{Zihao Zuo},
  \fpauthor[6]{Steven Dillmann},
  \fpauthor[2]{Yiyou Sun},
  \fpauthor[2]{Xinyang Han},\\
  \fpauthor[10]{Hanchung Lee},
  \fpauthor[12]{Molei Tao},
  \fpauthor[20]{Yue Zhao},
  \fpauthor[24]{Shilong Liu},
  \fpauthor[5]{Yuchen Wu},
  \fpauthor[2]{Xuandong Zhao},\\
\mbox{\fpauthor[1]{Xiangyi Li}, \fpauthor[37]{Florian Brand}, \fpauthor[2]{Gerbrand Ceder}, \fpauthor[2]{Hartmut Haeffner}}}
\affiliations{\fpaffil{1}{BenchFlow}, \fpaffil{2}{University of California, Berkeley}, \fpaffil{3}{California Institute of Technology}, \fpaffil{4}{Harvard University}, \fpaffil{5}{University of Washington}, \fpaffil{6}{Stanford University}, \fpaffil{7}{Amazon}, \fpaffil{8}{University of Pittsburgh}, \fpaffil{9}{University of California, San Diego}, \fpaffil{10}{Independent Researcher}, \fpaffil{11}{ZEISS}, \fpaffil{12}{Georgia Institute of Technology}, \fpaffil{13}{Lawrence Berkeley National Laboratory}, \fpaffil{14}{University of Michigan, Ann Arbor}, \fpaffil{15}{The Scripps Research Institute}, \fpaffil{16}{Oak Ridge National Laboratory}, \fpaffil{17}{University of Texas at Austin}, \fpaffil{18}{Purdue University}, \fpaffil{19}{University of Toronto}, \fpaffil{20}{University of Southern California}, \fpaffil{21}{Ludwig-Maximilians-Universit\"at M\"unchen}, \fpaffil{22}{Carnegie Mellon University}, \fpaffil{23}{University of California, Davis}, \fpaffil{24}{Princeton University}, \fpaffil{25}{Northeastern University}, \fpaffil{26}{University of Pennsylvania}, \fpaffil{27}{University of Minnesota, Twin Cities}, \fpaffil{28}{New York University}, \fpaffil{29}{Duke University}, \fpaffil{30}{Nanyang Technological University}, \fpaffil{31}{Pipeline Lab}, \fpaffil{32}{RLWRLD}, \fpaffil{33}{University of Palermo}, \fpaffil{34}{IBM}, \fpaffil{35}{University of Amsterdam}, \fpaffil{36}{University of California, Santa Cruz}, \fpaffil{37}{Prime Intellect}}
\titlenotes{\textsuperscript{*}Equal contribution. \textsuperscript{\dag}Core contribution. \textsuperscript{\S}Work conducted outside of the authors' roles at Amazon.}
\begin{document}

\maketitle
\suppressfloats[t]

\begin{abstract}
AI agents are playing an increasingly important role in physics research. However, existing physics benchmarks only evaluate fixed questions, isolated sub-tasks, or paper reproduction with given solutions, and rarely check research quality alongside correctness. We introduce \textsc{FrontierPhysics}, an end-to-end benchmark with 56 tasks made by 42 physicists based on their own research, covering nine domains across theoretical, computational, and experimental physics. Each task requires agents to review the literature, develop a research plan, conduct research and experiments, and produce structured outputs and scientific manuscripts. Combining deterministic verifiers with expert-authored rubrics, \textsc{FrontierPhysics} evaluates whether agents' research is correct and publishable. Across 22 models in \FPModelCount{} configurations, no configuration passes more than \FPOpusFiveFivePasses{} of its 168 rollouts. Manual review of agent trajectories shows that frontier agents’ main weaknesses are insufficiently rigorous reasoning and limited expert intuition, rather than gaps in physics knowledge or execution errors.
\end{abstract}

\par\vspace{-\parskip}\noindent\begin{minipage}{\linewidth}
  \centering
  \includegraphics[width=\linewidth]{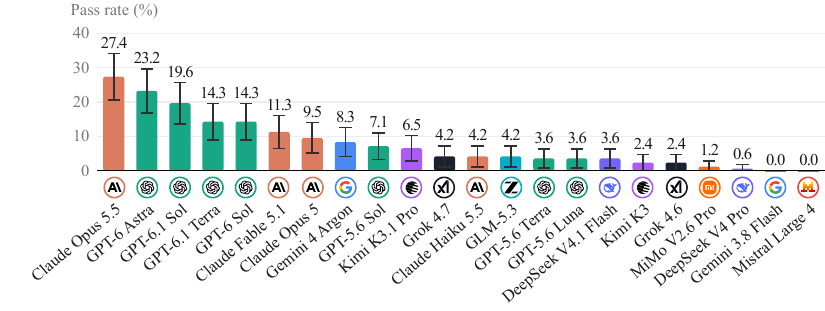}\par
  \vspace{-6pt}
  {\let\normalsize\small\captionof{figure}{\textbf{\textsc{FrontierPhysics leaderboard.}} 
Overall pass rate (Pass@1) under the Table~\ref{tab:main-results} configurations; a pass requires passing all verifier tests and blocker rubrics. Harnesses and inference level: GPT (Codex, max), Claude (Claude Code, max), Grok (Grok Build, xhigh), Gemini (Antigravity, high), Kimi (Claude Code, max), DeepSeek (Claude Code, max), MiMo (Claude Code, max), GLM (Claude Code, max), Mistral (Claude Code, max). Error bars show approximate 95\% confidence intervals.}\label{fig:teaser}}
\end{minipage}\par

\section{Introduction}

Large language models (LLMs) are evolving from passive scientific assistants into agents that take on parts of the research process. When equipped with external tools, these agents can search and summarize literature, formulate hypotheses, write and debug scientific code, and produce technical reports. Recent systems have demonstrated tool-augmented reasoning in chemistry, iterative code-based experimentation, multi-agent hypothesis generation, scientific-software optimization, and even interaction with laboratory instruments~\citep{bran2024chemcrow,boiko2023coscientist,lu2024aiscientist,schmidgall2025agentlaboratory,gottweis2026coscientist,aygun2026empiricalsoftware,novikov2025alphaevolve,swanson2025virtuallab,huang2025cascade,huang2025skillpuzzler,he2026xas,mandal2025afmagents,fei2026sdl}. These systems point toward agents that contribute to more complete scientific workflows and projects, not only isolated subtasks~\citep{chu2026awm}.

\begin{figure}[!t]
  \centering
  \includegraphics[width=\linewidth]{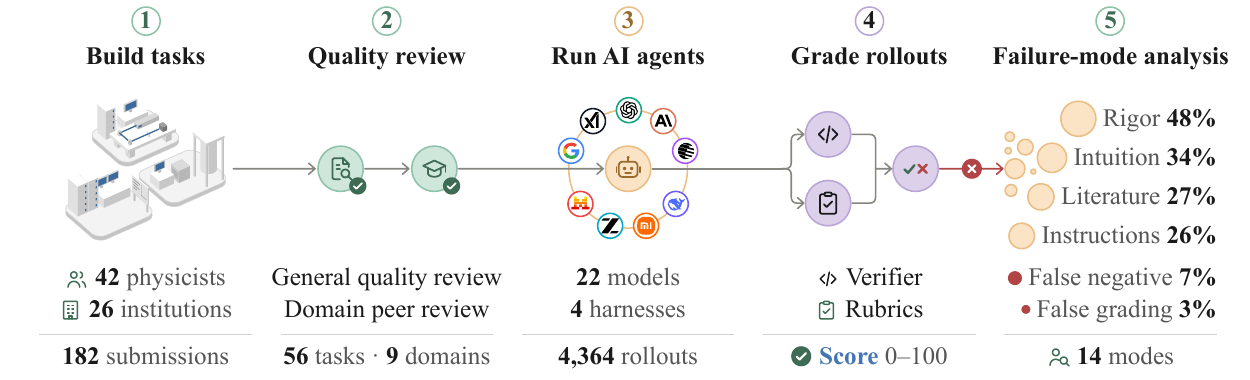}
  \vspace{-12pt}
  \caption{
\textbf{Overview of \textsc{FrontierPhysics}.} (1)~Physicists build tasks from their own research. (2)~An admin team of physicists accepts those passing both reviews. (3)~4,364 rollouts cover 22 models and 4 harnesses, each is one trial of a task by a model in an agent harness. (4)~A deterministic verifier and a reviewer agent with the task's rubric grade each rollout (Eq.~\ref{eq:score}). (5)~The admin team and contributors manually analyze failure modes of the 168 GPT-6 Astra rollouts (Fig.~\ref{fig:failure-modes}, Table~\ref{tab:failure-taxonomy}).}
\label{fig:benchmark-overview}
\end{figure}

Evaluating progress toward this goal requires benchmarks that capture the open-ended, multi-stage nature of scientific research. Scientific benchmarks are accordingly shifting from textbook-level questions~\citep{physics,olympicarena,ugphysics,phybench,cmphysbench} toward expert-level closed-ended questions~\citep{hle} and research-level problems~\citep{ale}. Among existing research-level benchmarks, some assess scientific reasoning~\citep{tpbench,critpt,cmt-benchmark,li2026rxnbench,wang2026epibench}, while others evaluate the agents' ability to complete individual tasks within realistic scientific workflows~\citep{frontierscience,realpdebench,mandal2025afmagents,stargazer,simulcost,sciskillbench_data}. However, these benchmarks target isolated capabilities, and because many of their tasks are closed-ended with deterministic answers, models quickly saturate them, so the benchmarks lose signal. To move toward end-to-end research, some benchmarks formulate tasks as full-paper reproductions~\citep{prbench,qmpbench,replicationbench,colliderbench}. Yet because agents are given the original publications, the research questions and solution structures are largely specified in advance, limiting their ability to evaluate open-ended scientific inquiry. More recent benchmarks address this limitation by converting published studies into research problems without revealing their solutions~\citep{prlbench,naturebench,researchclawbench,physcibench}. Nevertheless, designing and evaluating such tasks involves many subtleties that require the expertise of the original authors: for example, how to write instructions that preserve as much open-endedness as possible while avoiding false negatives, and how to determine if an agent's failure reflects true scientific misunderstanding or a lack of expert intuition and experience. To our knowledge, no existing benchmark is fully agentic, built by physicists from their own research, graded on both paper quality and result correctness, and designed to address these subtleties through manual task review and failure-mode analysis (Table~\ref{tab:physics-benchmarks}).

In this paper, we introduce \textbf{\textsc{FrontierPhysics}} (Fig.~\ref{fig:teaser}), a benchmark evaluating whether AI agents can conduct end-to-end research across theoretical, computational, and experimental physics.
\textsc{FrontierPhysics} tasks are designed by active physics researchers (including Ph.D. students, postdocs, and research scientists) from research projects they have personally conducted. On average, each task author has 420 citations. To preserve open-endedness while minimizing false negatives and positives, each task goes through manual general review, domain-expert peer review, and failure-mode analysis.
For each task (Fig.~\ref{fig:benchmark-overview}), an agent receives a high-level prompt that resembles the instructions provided by a principal investigator (PI) at the beginning of a research project. The agent must then carry out the research and produce both a scientific manuscript and numerical outputs that can be verified deterministically. Much of this process is open-ended: the agent must investigate the topic, formulate a research plan, select appropriate methods, validate the results, and establish the significance of its findings. To evaluate these capabilities, \textsc{FrontierPhysics} combines a deterministic verifier (test scripts that check the numerical outputs) with rubric-based grading by reviewer agents.

\begin{table*}[!b]
\centering
\caption{
\textbf{Comparison of scientific research benchmarks.}
We compare scientific research benchmarks across the research authenticity, research components covered in the evaluation, and the evaluation methods. Meanings of labels: \textbf{Tasks/Phys.}: total/physics task counts in released task lists. \textbf{Own Research}: the tasks come from the contributors' own research. \textbf{No Reference Given}: there is no reference (e.g., the original paper) provided to the agent. \textbf{Literature Graded}: the quality of agents' literature review and synthesis is evaluated. \textbf{Write-up Graded}: agents write up scientific manuscripts and the manuscripts are graded. \textbf{Expert Rubric}: at least part of the evaluation protocol uses rubrics created by domain experts. \textbf{Deterministic Verifiers}: at least part of the evaluation protocol contains deterministic verifiers. \textbf{Failure Analysis}: domain experts manually check the trajectories and output papers of selected rollouts.
$\checkmark$: documented; $\circ$: partial; $\times$: no.
}
\label{tab:physics-benchmarks}

\begingroup
\setlength{\tabcolsep}{3pt}
\renewcommand{\arraystretch}{1.08}

\resizebox{\textwidth}{!}{\begin{tabular}{@{}l*{9}{c}@{}}
\toprule
\textbf{Benchmark}
& \textbf{Tasks}
& \textbf{Phys.}
& \multicolumn{2}{c}{\textbf{Research Authenticity}}
& \multicolumn{2}{c}{\textbf{Research Coverage}}
& \multicolumn{3}{c}{\textbf{Evaluation}} \\
\cmidrule(lr){4-5}
\cmidrule(lr){6-7}
\cmidrule(lr){8-10}
& &
& \shortstack{\textbf{Own}\\\textbf{Research}}
& \shortstack{\textbf{No Reference}\\\textbf{Given}}
& \shortstack{\textbf{Literature}\\\textbf{Graded}}
& \shortstack{\textbf{Write-up}\\\textbf{Graded}}
& \shortstack{\textbf{Expert}\\\textbf{Rubric}}
& \shortstack{\textbf{Deterministic}\\\textbf{Verifiers}}
& \shortstack{\textbf{Failure}\\\textbf{Analysis}}\\
\midrule
\rowcolor{black!7}
\multicolumn{10}{@{}l}{\textit{Reasoning and QA}} \\
CritPt~\citep{critpt}
& 71 & 71
& $\checkmark$ & $\checkmark$ & $\times$
& $\times$ & $\times$ & $\checkmark$ & $\checkmark$ \\
FrontierScience-Research~\citep{frontierscience}
& 60 & 20
& $\checkmark$ & $\checkmark$ & $\times$
& $\times$ & $\checkmark$ & $\times$ & $\times$ \\

\midrule
\rowcolor{black!7}
\multicolumn{10}{@{}l}{\textit{Subtasks from scientific workflows}} \\
Terminal-Bench-Science~\citep{terminalbenchscience}
& 70 & 5
& $\checkmark$ & $\checkmark$ & $\times$
& $\times$ & $\times$ & $\checkmark$ & $\circ$ \\

\midrule
\rowcolor{black!7}
\multicolumn{10}{@{}l}{\textit{Paper reproduction}} \\
PRBench~\citep{prbench}
& 30 & 30
& $\circ$ & $\times$ & $\times$
& $\times$ & $\checkmark$ & $\times$ & $\circ$ \\
ReplicationBench~\citep{replicationbench}
& 111 & 111
& $\circ$ & $\circ$ & $\times$
& $\times$ & $\checkmark$ & $\checkmark$ & $\checkmark$ \\
QMP-Bench~\citep{qmpbench}
& 100 & 100
& $\times$ & $\times$ & $\times$
& $\times$ & \checkmark & \checkmark & $\circ$ \\

\midrule
\rowcolor{black!7}
\multicolumn{10}{@{}l}{\textit{End-to-end research}} \\
NatureBench~\citep{naturebench}
& 90 & 13
& $\times$ & $\checkmark$ & $\times$
& $\times$ & $\times$ & $\checkmark$ & $\circ$ \\
PRL-Bench~\citep{prlbench}
& 100 & 100
& $\times$ & $\checkmark$ & $\times$
& $\times$ & $\checkmark$ & $\checkmark$ & $\circ$ \\
ResearchClawBench~\citep{researchclawbench}
& 40 & 4
& $\times$ & $\checkmark$ & $\times$
& $\checkmark$ & $\checkmark$ & $\times$ & $\circ$ \\
PhySciBench~\citep{physcibench}
& 200 & 100
& $\times$ & $\checkmark$ & $\circ$
& $\times$ & $\checkmark$ & $\checkmark$ & $\circ$ \\

\midrule
\rowcolor{ours}
\textbf{\textsc{FrontierPhysics}}
& \textbf{56} & \textbf{56}
& $\boldsymbol{\checkmark}$ & $\boldsymbol{\checkmark}$
& $\boldsymbol{\checkmark}$
& $\boldsymbol{\checkmark}$ & $\boldsymbol{\checkmark}$
& $\boldsymbol{\checkmark}$
& $\boldsymbol{\checkmark}$\\
\bottomrule
\end{tabular}}
\endgroup
\end{table*}

\section{FrontierPhysics}
\label{sec:benchmark}

\begin{figure}[!t]
\centering
\includegraphics[width=\linewidth]{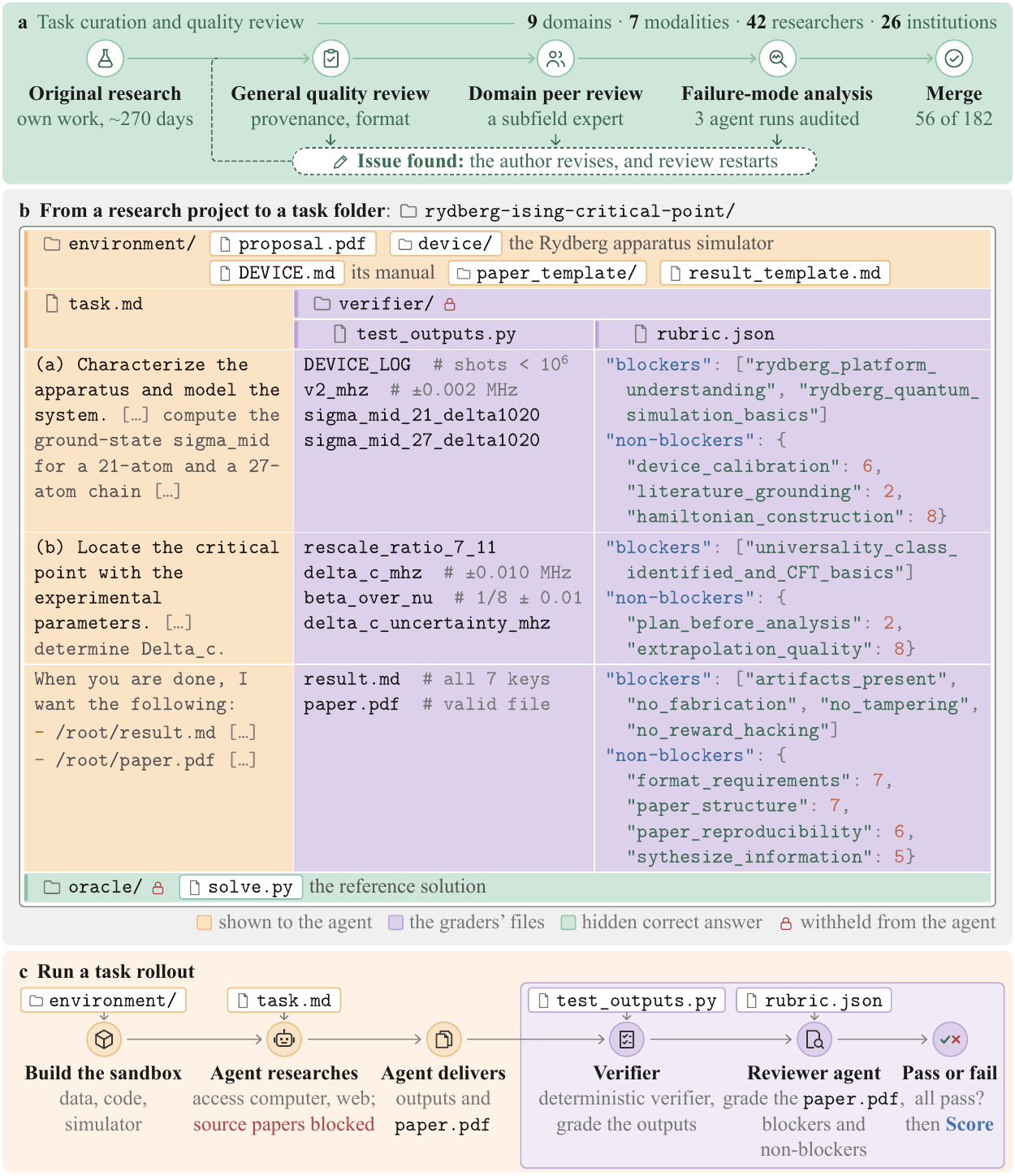}
\caption{(a), \textsc{FrontierPhysics} tasks QA pipeline. (b), \textsc{FrontierPhysics} tasks structure, shown for the example task \texttt{rydberg-\allowbreak ising-\allowbreak critical-\allowbreak point}. (c), How to run a task rollout.}
\label{fig:benchmark-details}
\end{figure}

Figure~\ref{fig:benchmark-details} shows how a \textsc{FrontierPhysics} task is curated (a), structured (b), and run (c).

\subsection{Task Design}
\label{task_design}
Tasks are designed to reconstruct end-to-end research projects. Figure~\ref{fig:benchmark-details}b shows an example task and Appendix~\ref{app:task-example} gives its full instruction and proposal. Some important questions to figure out or directions to explore might be given, but no explicit methodology or solution pathways will be specified. When necessary, input data and other supporting files are provided in the task environment (Fig.~\ref{fig:benchmark-details}b) to enable the agent to configure and execute the relevant computational or experimental workflow.

Each task requires two types of final deliverables: deterministic artifacts, such as computed quantities, simulation outputs, or processed data, and a paper manuscript submitted as \texttt{paper.pdf}. The paper is expected to meet the standards of a peer-reviewed paper by clearly presenting the research context, methodology, results, and conclusions, while faithfully incorporating the evidence and computational artifacts produced during task execution.

The rubric-based assessment includes global rubrics and task-specific rubrics. The global rubrics evaluate behavior alignment including reward hacking~\citep{amodei2016concrete,skalse2022defining},
hallucination and general research practices including literature synthesis, result validation, and the use of evidence to support scientific claims, whereas task-specific rubrics are developed by domain experts to assess the agent's understanding of the underlying physics, scientific presentation, and execution of the research task. Because rubric design requires task-specific expertise, we ask each contributor to design the rubrics for their own task and research project. Depending on the task, the rubric criteria may assess whether the agent identifies and correctly uses key references, discovers the key scientific ideas through its investigation, considers realistic experimental constraints, or provided good motivations for its work in the paper.
\newpage 
\subsection{Task Curation and Quality Review}
\label{task_QA}
\paragraph{General quality review.} When a contributor creates a task submission, we first assess its authenticity and provenance (Fig.~\ref{fig:benchmark-details}a): the contributor must provide supporting information (e.g., reference papers or patent links) to demonstrate that they have personally conducted the project. Contributors are asked to provide the project background, a timeline of the original research process, and the amount of time they spent completing it. We require the underlying project to have involved at least two weeks of research effort. We then check that \texttt{task.md} and \texttt{rubric.json} are human-readable and contain genuine human input, rejecting tasks that appear purely AI-generated.

\paragraph{Domain peer review.} We aim to have every task reviewed by an expert in a closely related field. To scale this, we invite contributors to become domain reviewers once their own tasks pass the first-stage general quality check. Our domain reviewers have therefore all conducted research in their fields, and consist mostly of Ph.D. students (47\%) and Ph.D. holders (37\%). Each new reviewer is onboarded through a one-on-one meeting, and we audit the quality of their first review to align standards across reviewers. When a domain reviewer judges a task ready to merge, we perform a final check against our standards.

Each domain reviewer is provided with a domain-review checklist which covers three aspects: First, the reviewer verifies that \texttt{task.md} does not over-specify the problem: the prompt should not contain domain knowledge that a competent researcher is expected to know or figure out, so that the task remains as open as possible. Second, the reviewer confirms that \texttt{rubric.json} encodes real physics insight that only someone who has conducted the research would possess, rather than the low-signal criteria typical of AI-generated rubrics. Third, the reviewer examines failure modes after getting agents trajectories: contributors must perform at least three local runs with SOTA agents and an initial failure-mode analysis, and the reviewer inspects these traces for finding potential false positives and false negatives. Common issues include unjustified numerical tolerances that fail agents implementing a correct physical model or ambiguous instructions that lead to agents' misunderstanding.

When all checks pass, an admin team with six physics PhD researchers from diverse subfield backgrounds approves and merges the submission into the final task set. In total, 56 tasks were accepted from 42 physics researchers from 26 institutions worldwide in 9 different physics domains out of 182 initial task submissions. A summary can be seen in Fig.~\ref{fig:task-composition}. For 18 tasks, the underlying research has been published in papers co-authored by the task contributors. These include four tasks based on papers published in Nature, with other journals including Nature Communications, Physical Review Letters, and Physical Review X. On average contributors report to have spent about 270 days per task for the original projects, and in total more than 8,000 hours of research.

\subsection{Evaluation}
\label{task_grading}
In each task rollout (Fig.~\ref{fig:benchmark-details}c), our evaluation scheme is designed to assess both the objective correctness of the agent's results and the overall quality of its research. To this end, we combine programmatic deterministic verifiers with an Agent-as-judge. The deterministic verifier checks structured outputs against reference values. After the agents finish running, agent trajectory files, verifier results, and output files including \texttt{paper.pdf} are recorded and saved. These files are then copied into another sandbox for a reviewer agent to review and grade the trajectory files and \texttt{paper.pdf} based on a \texttt{rubric.json}. We use an agent rather than a single LLM call as the judge because many rubric criteria, such as those on reward hacking, hallucinated results, and literature search, must be checked against the agent's trajectory, code, and output files rather than \texttt{paper.pdf} alone. All rollouts are graded by the same reviewer agent, GPT-5.6 Sol (xhigh) in Codex CLI. Section~\ref{sec:reubric_reliability} checks its grades against an expert audit (Table~\ref{tab:failure-tasks}, Fig.~\ref{fig:failure-modes}) and on the authors' own papers (Table~\ref{tab:reference-material-evaluation}). Regrading rollouts of the first evaluation round with Claude Opus 5.5 in Claude Code CLI as the reviewer agent gives the same pass/fail decision for almost all GPT-6 Astra rollouts, but the Opus 5.5 reviewer is more lenient on rollouts from its own model family (Table~\ref{tab:reviewer-regrading}). Workspace files and trajectory files of the reviewer agents are collected the same way to HuggingFace datasets for later analysis.

Rubric-based grading has been widely adopted in benchmarks involving open-ended exploration or qualitative dimensions of agent performance~\citep{paperbench,healthbench,sedareval,prbench,frontiercode11}. Whereas deterministic verification establishes whether an output is correct, rubric-based evaluation assesses more aspects, for example whether the agent conducted and communicated the research well. Inspired by the rubric design of FrontierCode~\citep{frontiercode11}, we divide our rubric criteria into \emph{blockers} and \emph{non-blockers}. Blockers are binary gates that represent requirements essential to the scientific validity of a task. Thus, an agent must satisfy every blocker as well as the deterministic verifier to pass. Non-blockers capture qualitative aspects of the work. Each non-blocker is scored on a three-level scale: 0 indicates that the requirement is absent or incorrect, 1 indicates partial satisfaction, and 2 indicates complete satisfaction. The precise anchors for these levels are defined separately for each criterion. The task author also gives every criterion an integer \emph{weight} from 1 to 10 for its relative importance. Blocker weights are retained as rubric metadata but do not contribute to the numerical quality score. The final score of a task:
\begin{equation}
S_{\mathrm{final}}
=
\begin{cases}
\displaystyle
\frac{\sum_{i \in \mathcal{N}} w_i s_i}
     {2\sum_{i \in \mathcal{N}} w_i}\times 100,
& \text{if } D=1 \text{ and } b_j=1 \;\; \forall j \in \mathcal{B}, \\[10pt]
0, & \text{otherwise},
\end{cases}
\label{eq:score}
\end{equation}
Here, $\mathcal{N}$ and $\mathcal{B}$ denote non-blockers and blockers respectively, $s_i \in \{0, 1, 2 \}$, $w_i$ is the criterion weight, $D$ is the deterministic verifier result, and $b_j$ is the blocker result.

$S_{\mathrm{final}}=0$ indicates that the results are not publishable, while the value of $S_{\mathrm{final}}$ when $S_{\mathrm{final}} \neq 0$ measures the quality of the publication. High scores on non-blockers cannot compensate for an incorrect deterministic result or a failed blocker. Conversely, an agent may pass all deterministic checks while producing a poorly motivated, inadequately supported, or badly communicated manuscript. Such a submission passes the verifiers and blockers but receives a low $S_{\mathrm{final}}$.

\section{Experiments}
\label{sec:experiments}

We evaluate 22 frontier models in four agent harnesses. Each model attempts all 56 tasks three times at its highest available inference effort, and four GPT models are also run at xhigh.

\subsection{Experimental Setup}

\label{sec:experimental-setup}

As in SkillsBench \cite{skillsbench} and ClawsBench \cite{clawsbench} we use BenchFlow CLI~\cite{benchflow}, a research infra command line tool for creating RL environments, post-training, and evals, to run experiments as well as track agent trajectories, verifier, and rubrics results. We use Codex CLI for GPT models, Claude Code CLI for Claude models, Grok build for Grok models, and Antigravity CLI for Gemini models. The other seven models, DeepSeek V4 Pro, DeepSeek V4.1 Flash, GLM-5.3, Kimi K3, Kimi K3.1 Pro, MiMo V2.6 Pro and Mistral Large 4, also run in Claude Code CLI. Each agent runtime is started in parallel cloud sandboxes. For tasks with resource requirements under 4 vCPUs, 16GB RAM, and 10GB storage, we use Daytona as a sandbox provider. For tasks with higher resource requirements, we use Docker in Azure VMs with Lambda API access so the model can call GPU resources as needed.

All results are for \textsc{FrontierPhysics} v0.1, evaluated on all 56 tasks. Where a task changed after the first evaluation round in a way that could alter a result, the affected rollouts were re-run or regraded; the other changes were judged unable to alter a result.

Each rollout has a uniform 10-hour timeout cap. Endings caused by the model, such as reaching the budget or exhausting the sandbox's memory, are final and stay in the denominator; endings caused by the infrastructure, such as a lost sandbox or provider errors that outlast retries, are re-run. Table~\ref{tab:zero-filled} counts the slots left without a valid result.

\subsection{Results}
\label{sec:model-performance}
\begin{figure}[!t]
\centering
\includegraphics[width=\linewidth]{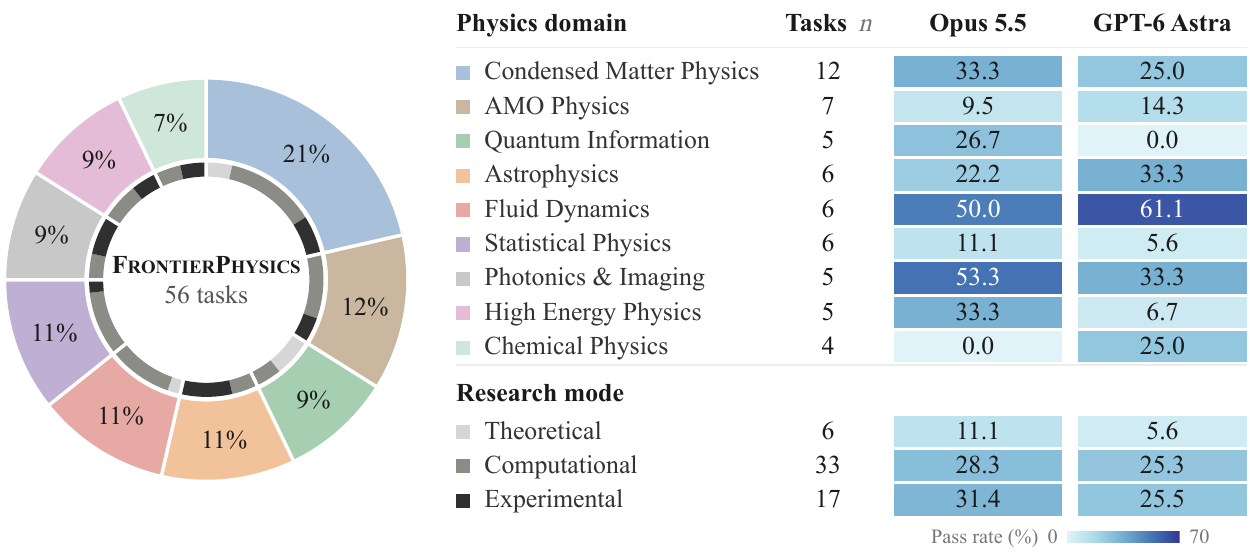}
\caption{Task composition and overall pass rate (\%) by primary physics domain across all \FPTaskCount{} tasks. Donut labels give task shares. $n$ gives task counts.}
\label{fig:task-composition}
\label{fig:subfield-performance}
\end{figure}

\begin{figure}[!t]
\centering
\includegraphics[width=\linewidth]{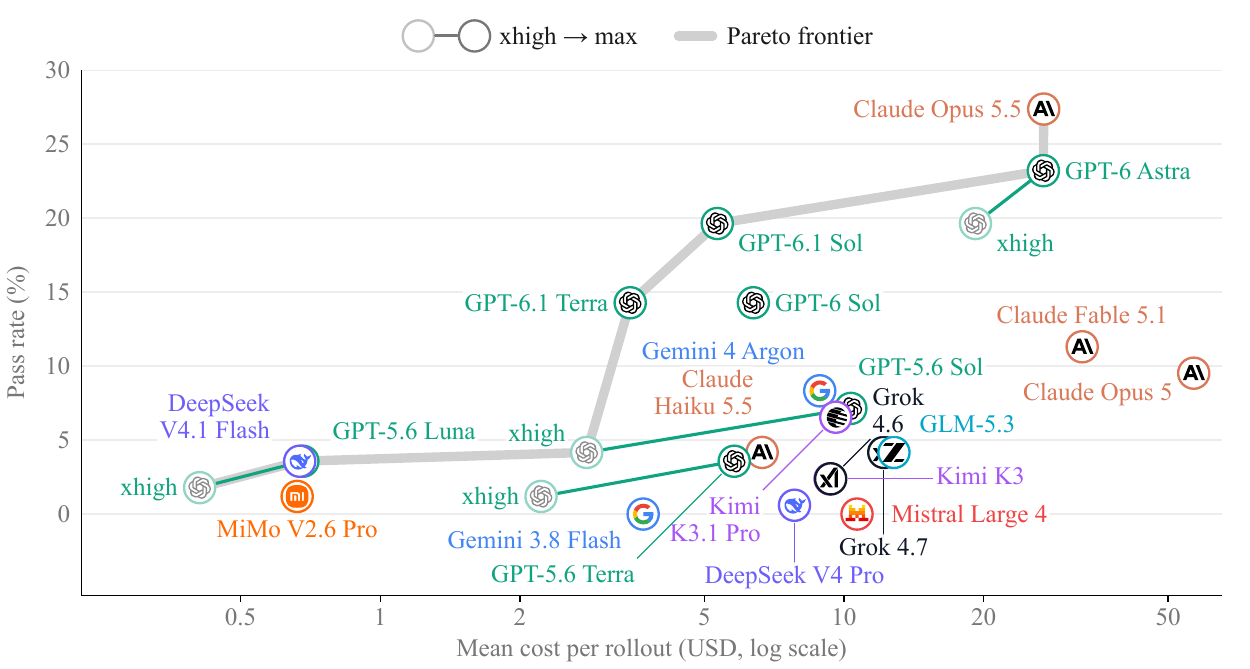}\par\medskip
\includegraphics[width=\linewidth,trim=0 0 0 27,clip]{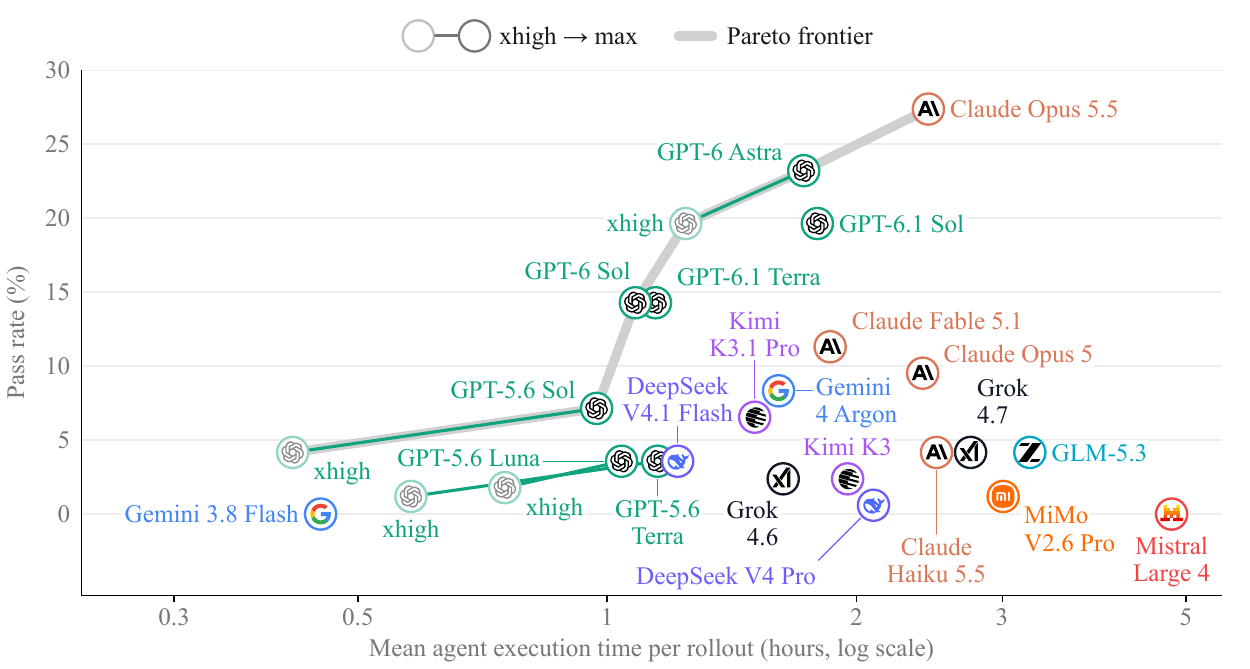}
\caption{Mean reported inference cost (top) and agent execution time (bottom) per rollout versus overall pass rate (\%). Time excludes environment setup, verification and rubric review.}
\label{fig:pareto-frontier}
\label{fig:agent-time}
\end{figure}

The result (Fig.~\ref{fig:teaser}) indicates that the overall pass rates are below 30\% for all agents tested. Claude Opus 5.5 has the highest pass rate, \FPOpusFiveFiveOverall{}. Table~\ref{tab:main-results} provides more detailed information on the different runs and scores.

Figure~\ref{fig:pareto-frontier} plots the pass rate versus the mean cost and time per rollout. Pass rate generally rises with cost. However, for some Claude models, the pass rate decreases as cost increases, in the order Opus 5.5, Fable 5.1, and Opus 5. In addition, GPT models are generally more cost-efficient. Figure~\ref{fig:subfield-performance} gives the pass rate per physics domain of Opus 5.5 and GPT-6 Astra; with few tasks per domain, the differences between domains are uncertain. Appendix~\ref{app:per-task-results} gives per-task results.

Figure~\ref{fig:agent-time} and Appendix~\ref{sec:agent-resources} compare agent time, steps and output tokens with the pass rate.
Table~\ref{tab:resource-stats} gives the mean, median and 95th percentile of tool calls, agent time and cost per rollout for each configuration, and the mean and median of output tokens.

\begin{table}[!t]
\centering
\caption{Detailed Frontier-model performance across all \FPTaskCount{} tasks. For the experiment we use the highest available inference effort by default, where configurable. GPT models also appear at xhigh. The Score column reports the score defined in Equation~\ref{eq:score}, averaged over the three rollouts for each task and then across all tasks. Deterministic pass requires all verifier tests to pass and overall pass additionally requires all blocking rubric criteria to pass. Pass@1 gives the average pass rate over the three rollouts, while Pass@3 gives the percentage of tasks with at least one passing rollout. Uncertainties are approximate 95\% binomial confidence intervals over all rollouts, as in Figure~\ref{fig:teaser}. Table~\ref{tab:ci-main-results} gives 95\% confidence intervals from a bootstrap over tasks, which account for the correlation of the three trials of a task. Rows are ordered by overall Pass@1, then by deterministic Pass@1 and Score.}
\label{tab:main-results}
\includegraphics[width=\linewidth]{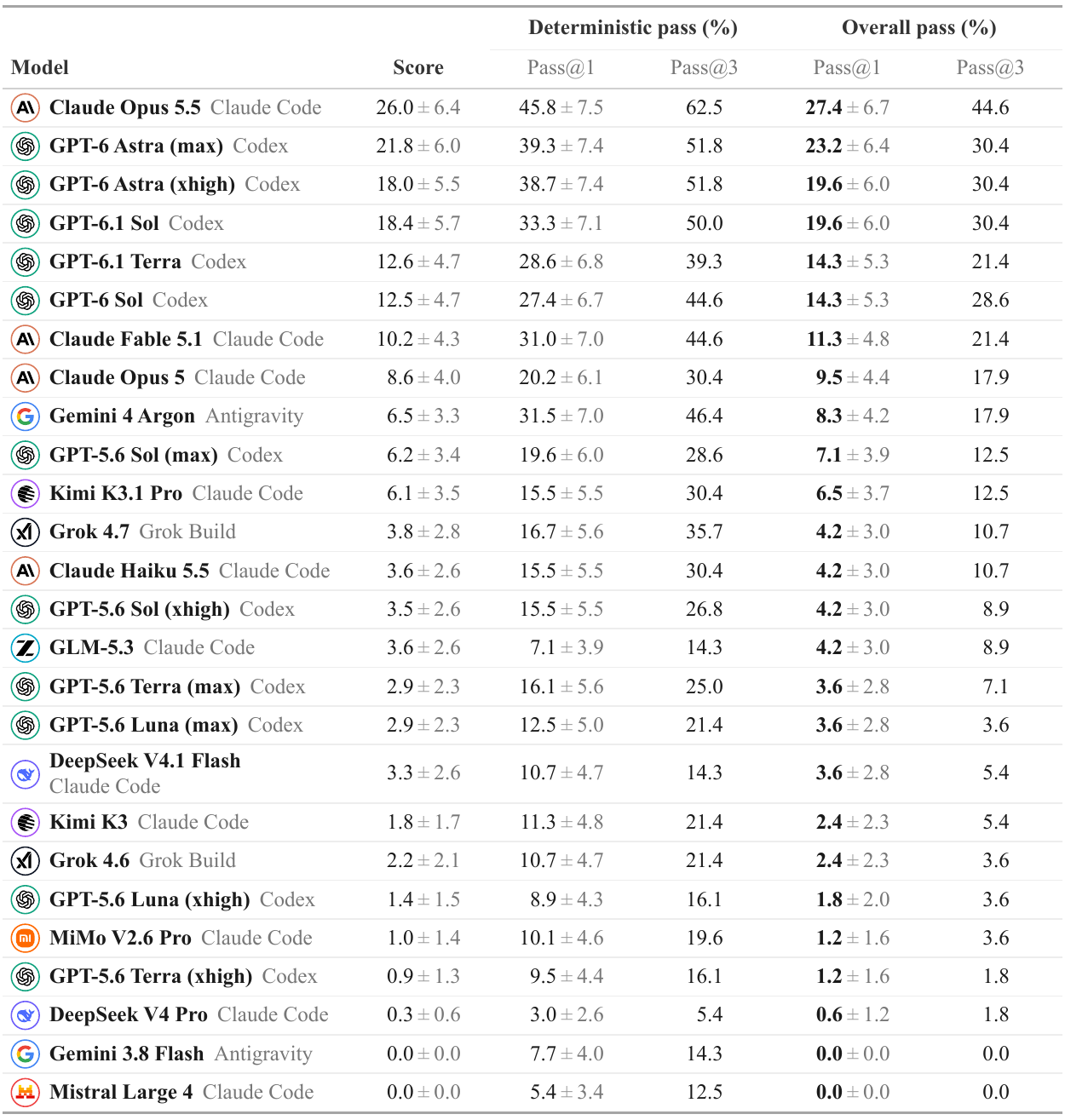}
\end{table}

\subsection{Assessing Verifier and Rubric Reliability}
\label{sec:reubric_reliability}
Verifier and rubric defects may remain after review. One major source of defects is adversarial verifier or rubric engineering designed to suppress task pass rates, such that even correct solutions produced by human experts may fail. To mitigate this, we have a three stage process:

First, we ask every task contributor to do a failure-mode report of the three GPT-6 Astra rollouts, as this is the best performing model with single agent harness when the analysis started, check the agent trajectories and output files, and explain where the agent failed and why. Second, our admin team reviewers re-read these contributors' failure-mode reports and double check them against the agent trajectories and output files. After reviewing around 60 rollouts, reviewers consolidated the recurring failures into the taxonomy in Table~\ref{tab:failure-taxonomy}. False negatives, where agents deliver the correct results but failed by verifier or reviewer agent, are found during this process and are labeled in Appendix~\ref{app:failure-labeling}. As highlighted in Table~\ref{tab:failure-tasks}, rollouts with exclusively false failures lead to an underestimated pass rate. In total, nine of the 168 rollouts incorrectly failed.
This coincides with the recent research \cite{re_grading}, in which domain experts regrade some leading benchmarks and find that many failures are due to grader errors, incorrect reference solutions, and ambiguous or underspecified questions. This suggests that finding false negatives takes both domain expertise and time, but is crucial for maintaining high benchmark quality.

The non-blocking rubric scores are harder to assess because they involve more subjective judgments. Third, as a sanity check of the rubric quality we asked authors to include their own papers or manuscripts as reference materials in the oracle when possible. Thirty-three tasks included these materials. For those with available reference, we selected the manuscript-level rubric criteria used for grading these materials, including blocker and non-blocker scoring criteria against the authors' reference work. We then use the same configuration as main experiment to run reviewer agents in sandboxes to grade these files with rubrics. The authors' own materials score 96.2\% on average under corresponding rubrics (Table~\ref{tab:reference-material-evaluation}), showing good alignment between rubrics grading with the human expert materials.

\subsection{Failure Mode Analysis}
\label{sec:failure-modes}

\begin{figure}[ht]
\centering
\vspace{-5pt}
\includegraphics[width=\linewidth]{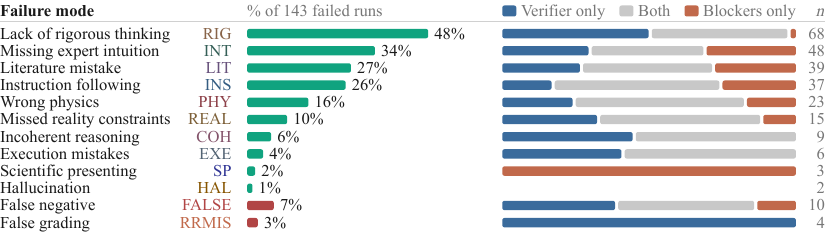}
\vspace{-16pt}
\caption{Failure mode analysis of GPT-6 Astra xhigh across 56 tasks (3 rollouts for each task). Counts $n$ and percentages ($n/143$) use the final review labels in Table~\ref{tab:failure-tasks}. Each rollout may have multiple tags. Verifier/rubric proportions are retained from the original recorded assessment. Correcting the eight FALSE and one RRMIS outcomes changes the pass rate from 14.9\% (25/168) to 20.2\% (34/168).}
\label{fig:failure-modes}
\end{figure}

To find why agents fail and how well the agent score approximates what we want to measure, we follow the protocol of Section~\ref{sec:reubric_reliability}. Figure~\ref{fig:failure-modes} summarizes the earlier matched-cohort annotations for all 168 single agent GPT-6 Astra rollouts. Table~\ref{tab:failure-tasks} reports the final review-sheet labels. More detailed information and an explanation of the failure modes can be found in Table~\ref{tab:failure-taxonomy}. This process found design issues in some tasks. We fixed them and re-ran or regraded the affected results (Section~\ref{sec:experimental-setup}). The pass rate of GPT-6 Astra (xhigh) after the re-runs (Table~\ref{tab:main-results}) agrees well with the manually corrected pass rate in Figure~\ref{fig:failure-modes}.

\begin{figure}[!t]
\centering
\includegraphics[width=\linewidth]{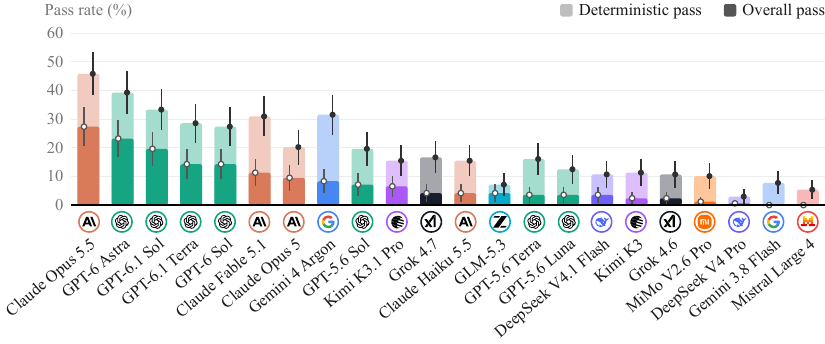}
\caption{Deterministic and overall pass rates (Pass@1, \%) of the models in Fig.~\ref{fig:teaser}. Column height: deterministic pass rate (all verifier tests pass). Solid part: overall pass rate (blocker rubrics also pass). Error bars: approximate 95\% confidence intervals (dot: deterministic; circle: overall).}
\label{fig:verifier-overall}
\end{figure}

Figure~\ref{fig:failure-modes} shows that agents mostly do not fail because of physics mistakes (e.g., a wrong calculation, misunderstanding of the physics concepts or lacking of physics knowledge), contrary to what one might expect from a physics benchmark. Instead, they often struggle to reason rigorously (48\%), apply good physical intuition when making decisions (34\%), and synthesize the literature (27\%). Typical rigorous thinking failures are: circular self-verification, where the agent's own validation method depends on its wrong assumption, so it can never find out its mistakes; and edge cases, where the agent fails to consider all the edge cases of its method or code, so that it fails at some hidden tests. The third most frequent failure is in the literature. In many cases, agents fail to find the key literature for the research (as identified by the task author), struggle to synthesize information from its literature search and provide a motivation for its project, or struggle to support its argument in the manuscript with literature. Missing physics intuition is another major failure mode. Typical examples of this failure mode include: agents fail to determine the major error source of an experiment, which is not because they do not understand it, but because they do not have the expert intuition to identify what is important; when debugging, agents cannot realize the root of the problem while keeping tuning the wrong knob. Besides the top three failure modes, instruction following accounts for 26\% of the failures. It could be agents missing some requirements in the instructions or misunderstanding the instructions. In many cases, the task authors believe the meaning should be clear to human domain experts, while the agent cannot get the explicit/implicit message.

The current failure mode analysis focuses on the GPT-6 Astra model (see Section~\ref{sec:reubric_reliability}) since it is challenging for tasks authors to manually read trajectory files with subagents involved. Other models, or other model-harness pairs, may fail in different ways, including ways outside this taxonomy. Also because of this, we can only accurately tag the false negatives of GPT-6 Astra's rollouts and therefore correct the final pass rate of GPT-6 Astra.

\section{Conclusion}
\label{sec:conclusion}

We have introduced \textsc{FrontierPhysics}, a benchmark dataset with 56 tasks evaluating the ability of AI agents to conduct physics research across nine domains. \textsc{FrontierPhysics} shows end-to-end physics research remains challenging to current models, given that Opus 5.5 in Claude Code CLI, the configuration with the highest pass rate, passes only \FPOpusFiveFivePasses{} of its 168 rollouts. Based on the human experts auditing of 168 SOTA agents rollouts trajectory and output files, we build a failure mode taxonomy and label all these 168 rollouts manually. From the taxonomy distribution, we conclude that lack of rigorous thinking (for example, circular self-verification) followed by missing expert intuition, instead of insufficient physics knowledge or execution mistakes, are the main weaknesses of frontier LLM agents.

Capable physics research agents should go beyond specified tasks to explore scientific problems independently, from problem definition to novel findings, with only high-level guidance. Training and evaluating such agents will require more open-ended research tasks and flexible evaluation frameworks that can assess exploratory reasoning, partial progress, and unexpected findings.

\bibliographystyle{frontierphysics}

\clearpage
\appendix
\raggedbottom
\pdfbookmark[0]{Appendix}{appendix-start}
\section*{Appendix}
\section{Example Task}
\label{app:task-example}
{\setlength\intextsep{0pt}\begin{figure}[H]
\centering
\includegraphics[width=\linewidth]{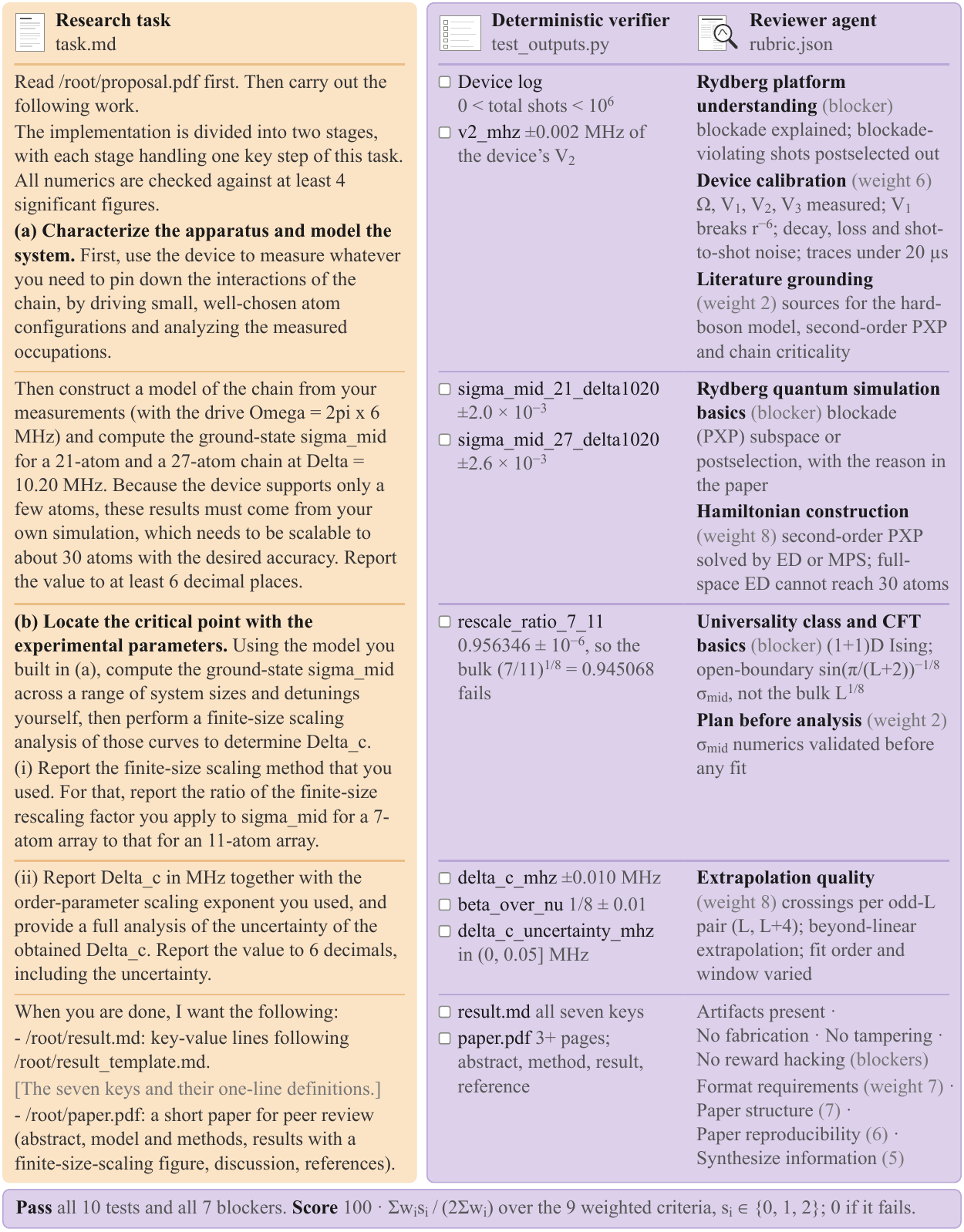}
\caption{\textbf{Example task.} \texttt{rydberg-\allowbreak ising-\allowbreak critical-\allowbreak point}, the example of Table~\ref{tab:failure-taxonomy}. Left: its \texttt{task.md}. The cited proposal (Figure~\ref{fig:task-proposal}) defines $\sigma_{\mathrm{mid}}$ as the order parameter on the central bond; the device simulates a Rydberg apparatus of at most three atoms. The tests that check each step in \texttt{task.md} ($\pm$: tolerance around withheld reference values) and the summarized rubric criteria that grade it.}
\label{fig:task-example}
\end{figure}}

\begin{figure}[H]
\centering
\includegraphics[width=\linewidth]{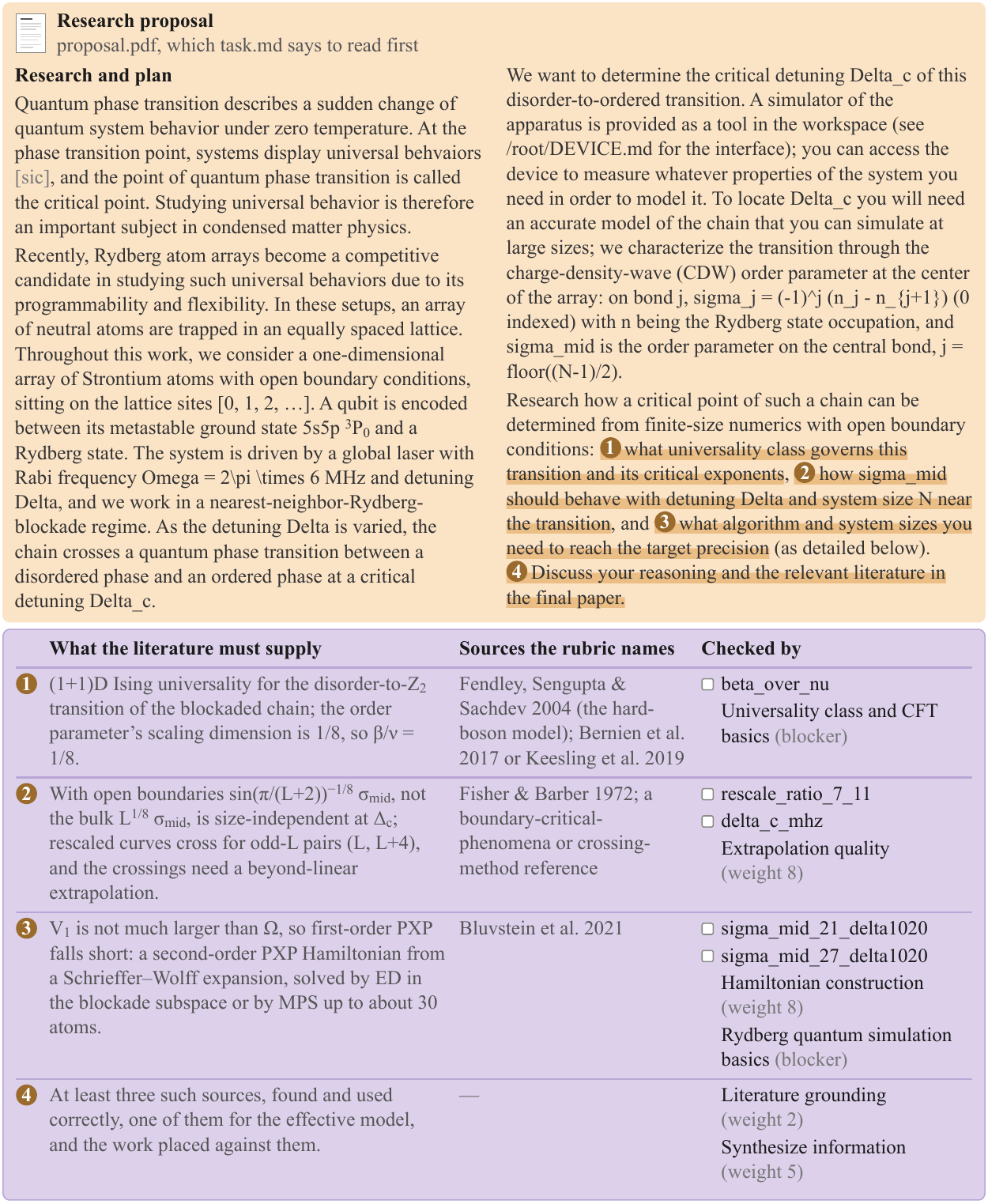}
\caption{\textbf{The proposal of the example task.} Top: \texttt{proposal.pdf}, the one-page brief that agent need to read first. It sets out the physics and the goal, and describe what are the questions should be answered after literature research. The last paragraph leaves four questions to the agent's own research (numbered). Bottom: for each question, what the literature must supply.}
\label{fig:task-proposal}
\end{figure}
 \section{Task Domain Taxonomy}
\label{app:task-domains}
The 56 tasks are assigned to nine primary domains. Each task contributes to exactly one primary-domain. The ``Also spans'' column in Table~\ref{tab:task-domains} records secondary domains for 15 tasks but these labels do not add extra observations to domain summaries.

\begingroup
\small
\setlength{\tabcolsep}{3pt}
\setlength{\LTleft}{0pt}
\setlength{\LTright}{0pt}
\setlength{\LTcapwidth}{\textwidth}
\renewcommand{\arraystretch}{1.3}
\begin{longtable}{@{}>{\raggedleft\arraybackslash}p{0.03\textwidth}>{\raggedright\arraybackslash}p{0.425\textwidth}>{\raggedright\arraybackslash}p{0.27\textwidth}>{\raggedright\arraybackslash}p{\dimexpr 0.275\textwidth-6\tabcolsep\relax}@{}}
\caption{Primary and secondary domain labels for all 56 \textsc{FrontierPhysics} tasks. A dash indicates no secondary domain label.}
\phantomsection\label{tab:task-domains}\\
\toprule
\textbf{\#} & \textbf{Task} & \textbf{Domain} & \textbf{Also spans} \\
\midrule
\endfirsthead
\caption[]{Task domain labels (continued).}\\
\toprule
\textbf{\#} & \textbf{Task} & \textbf{Domain} & \textbf{Also spans} \\
\midrule
\endhead
\midrule
\multicolumn{4}{r}{\small Continued on next page}\\
\endfoot
\bottomrule
\endlastfoot
1 & \texttt{arpes-\allowbreak sample-\allowbreak alignment} & Condensed Matter Physics & Photonics \& Imaging \\
2 & \texttt{baf2-\allowbreak ih001-\allowbreak strain-\allowbreak energy} & Condensed Matter Physics & - \\
3 & \texttt{cpt-\allowbreak frustrated-\allowbreak chain-\allowbreak spectra} & Condensed Matter Physics & - \\
4 & \texttt{eigenstate-\allowbreak optical-\allowbreak response-\allowbreak derivation} & Condensed Matter Physics & Photonics \& Imaging \\
5 & \texttt{lattice-\allowbreak band-\allowbreak geometry-\allowbreak discovery} & Condensed Matter Physics & - \\
6 & \texttt{li-\allowbreak graphene-\allowbreak mlip-\allowbreak transferability} & Condensed Matter Physics & Chemical Physics \\
7 & \texttt{llzo-\allowbreak proton-\allowbreak exchange-\allowbreak thermodynamics} & Condensed Matter Physics & Chemical Physics \\
8 & \texttt{lpbf-\allowbreak melt-\allowbreak pool-\allowbreak window} & Condensed Matter Physics & Fluid Dynamics \\
9 & \texttt{mlg-\allowbreak quantum-\allowbreak hall-\allowbreak dot-\allowbreak hartree-\allowbreak fock} & Condensed Matter Physics & - \\
10 & \texttt{moire-\allowbreak continuum-\allowbreak model-\allowbreak construction} & Condensed Matter Physics & - \\
11 & \texttt{rixs-\allowbreak entanglement-\allowbreak witness} & Condensed Matter Physics & - \\
12 & \texttt{smt-\allowbreak reflow-\allowbreak window} & Condensed Matter Physics & - \\
13 & \texttt{ion-\allowbreak trap-\allowbreak 3d-\allowbreak vs-\allowbreak 2d} & AMO Physics & - \\
14 & \texttt{ion-\allowbreak trap-\allowbreak design} & AMO Physics & - \\
15 & \texttt{multiplexing-\allowbreak ion-\allowbreak chain-\allowbreak qnet} & AMO Physics & Quantum Information \\
16 & \texttt{phase-\allowbreak controlled-\allowbreak optical-\allowbreak tweezer-\allowbreak holography} & AMO Physics & - \\
17 & \texttt{rydberg-\allowbreak cft-\allowbreak modulation-\allowbreak spectroscopy} & AMO Physics & - \\
18 & \texttt{rydberg-\allowbreak cz-\allowbreak phase-\allowbreak noise-\allowbreak infidelity} & AMO Physics & Quantum Information \\
19 & \texttt{rydberg-\allowbreak ising-\allowbreak critical-\allowbreak point} & AMO Physics & - \\
20 & \texttt{QNTK-\allowbreak research} & Quantum Information & - \\
21 & \texttt{cvqkd-\allowbreak calibrated-\allowbreak detector-\allowbreak noise} & Quantum Information & - \\
22 & \texttt{ising-\allowbreak qml-\allowbreak dwave} & Quantum Information & - \\
23 & \texttt{time-\allowbreak bin-\allowbreak qkd-\allowbreak product-\allowbreak povm} & Quantum Information & - \\
24 & \texttt{transmon-\allowbreak cz-\allowbreak pulse-\allowbreak engineering} & Quantum Information & - \\
25 & \texttt{desi-\allowbreak bao-\allowbreak sn-\allowbreak dark-\allowbreak energy} & Astrophysics & - \\
26 & \texttt{ems-\allowbreak scalar-\allowbreak spectral-\allowbreak census} & Astrophysics & - \\
27 & \texttt{nonlinear-\allowbreak boson-\allowbreak star-\allowbreak sideband-\allowbreak threshold} & Astrophysics & - \\
28 & \texttt{ns-\allowbreak eos-\allowbreak multimessenger} & Astrophysics & - \\
29 & \texttt{pulsar-\allowbreak timing-\allowbreak phase-\allowbreak connection} & Astrophysics & - \\
30 & \texttt{solar-\allowbreak wind-\allowbreak turbulence-\allowbreak spectral-\allowbreak break} & Astrophysics & - \\
31 & \texttt{consistent-\allowbreak two-\allowbreak phase-\allowbreak ls-\allowbreak vof} & Fluid Dynamics & - \\
32 & \texttt{diffuse-\allowbreak ib-\allowbreak rigid-\allowbreak body-\allowbreak fsi} & Fluid Dynamics & - \\
33 & \texttt{iga-\allowbreak branched-\allowbreak incompressible-\allowbreak flow} & Fluid Dynamics & - \\
34 & \texttt{material-\allowbreak agnostic-\allowbreak visual-\allowbreak inverse-\allowbreak dynamics} & Fluid Dynamics & - \\
35 & \texttt{multi-\allowbreak object-\allowbreak visual-\allowbreak inverse-\allowbreak dynamics} & Fluid Dynamics & - \\
36 & \texttt{twisted-\allowbreak slinky-\allowbreak wave-\allowbreak coupling} & Fluid Dynamics & - \\
37 & \texttt{glass-\allowbreak nonequilibrium-\allowbreak landscape-\allowbreak evolution} & Statistical Physics & - \\
38 & \texttt{kinked-\allowbreak ladder-\allowbreak polymer-\allowbreak scattering} & Statistical Physics & Chemical Physics \\
39 & \texttt{learning-\allowbreak credit-\allowbreak assignment} & Statistical Physics & - \\
40 & \texttt{low-\allowbreak dimensional-\allowbreak deep-\allowbreak network-\allowbreak learning} & Statistical Physics & - \\
41 & \texttt{metallic-\allowbreak glass-\allowbreak pel-\allowbreak ruggedness} & Statistical Physics & - \\
42 & \texttt{quantized-\allowbreak ising-\allowbreak 3sat-\allowbreak encoding} & Statistical Physics & Quantum Information \\
43 & \texttt{calibrated-\allowbreak cbct-\allowbreak local-\allowbreak reconstruction} & Photonics \& Imaging & - \\
44 & \texttt{fbg-\allowbreak rail-\allowbreak axle-\allowbreak inversion} & Photonics \& Imaging & - \\
45 & \texttt{fbg-\allowbreak strain-\allowbreak amplifier-\allowbreak design} & Photonics \& Imaging & - \\
46 & \texttt{heat-\allowbreak assisted-\allowbreak detection-\allowbreak and-\allowbreak ranging} & Photonics \& Imaging & - \\
47 & \texttt{hspic-\allowbreak spectrometer-\allowbreak e2e} & Photonics \& Imaging & - \\
48 & \texttt{corsika-\allowbreak air-\allowbreak shower-\allowbreak inverse} & High Energy Physics & Astrophysics \\
49 & \texttt{icecube-\allowbreak pointsource-\allowbreak unbinned} & High Energy Physics & Astrophysics \\
50 & \texttt{o3-\allowbreak monodromy-\allowbreak line-\allowbreak defect-\allowbreak mc} & High Energy Physics & Statistical Physics \\
51 & \texttt{o3-\allowbreak surface-\allowbreak cft-\allowbreak fuzzy-\allowbreak sphere} & High Energy Physics & Statistical Physics \\
52 & \texttt{sipm-\allowbreak spe-\allowbreak gain-\allowbreak breakdown} & High Energy Physics & Photonics \& Imaging \\
53 & \texttt{agi-\allowbreak surface-\allowbreak hydrophobicity} & Chemical Physics & - \\
54 & \texttt{hole-\allowbreak bifurcation-\allowbreak ml-\allowbreak landscape} & Chemical Physics & - \\
55 & \texttt{p2d-\allowbreak cell-\allowbreak model-\allowbreak identification} & Chemical Physics & - \\
56 & \texttt{smfret-\allowbreak research-\allowbreak v5} & Chemical Physics & Photonics \& Imaging \\
\end{longtable}
\endgroup
 \clearpage
\section{Input Modalities and Research Activities}
\label{app:task-characteristics}

\begin{figure}[htbp]
\centering
\includegraphics[width=\linewidth]{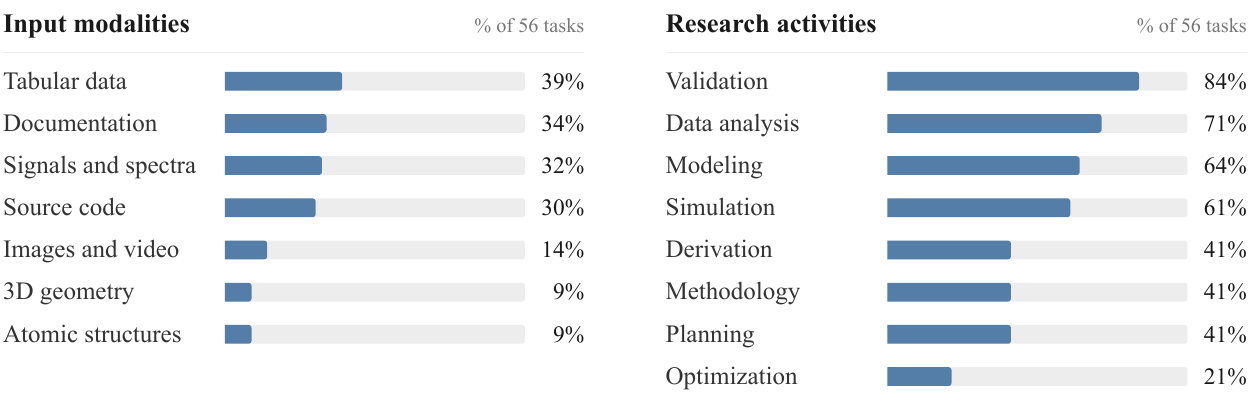}
\caption{Input modalities and required research activities across the 56 accepted tasks. Each bar gives the percentage of tasks carrying that label. Modality and activity categories are non-exclusive: a single task can contribute to multiple categories, so the percentages within each panel need not sum to 100\%.}
\label{fig:task-characteristics}
\end{figure}
 \section{Per-Task Results}
\label{app:per-task-results}
\begin{figure}[p]
\centering
\includegraphics[width=\linewidth]{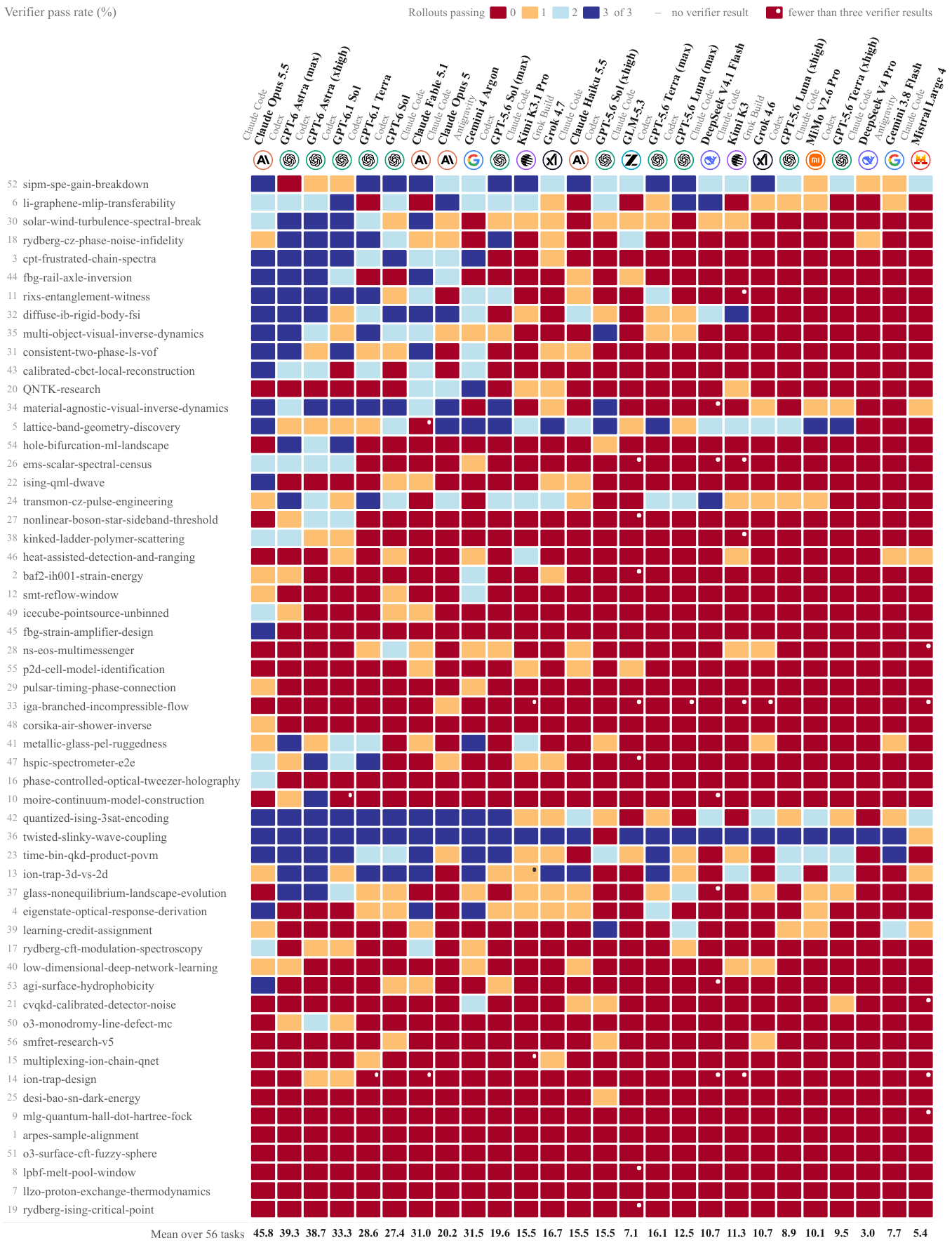}
\caption{Verifier pass rate per task and model. The bottom row is each model's deterministic Pass@1 over all \FPTaskCount{} tasks (Table~\ref{tab:main-results}). Unavailable outcomes contribute zero to the aggregate.}
\label{fig:per-task-verifier}
\end{figure}

\begin{figure}[p]
\centering
\includegraphics[width=\linewidth]{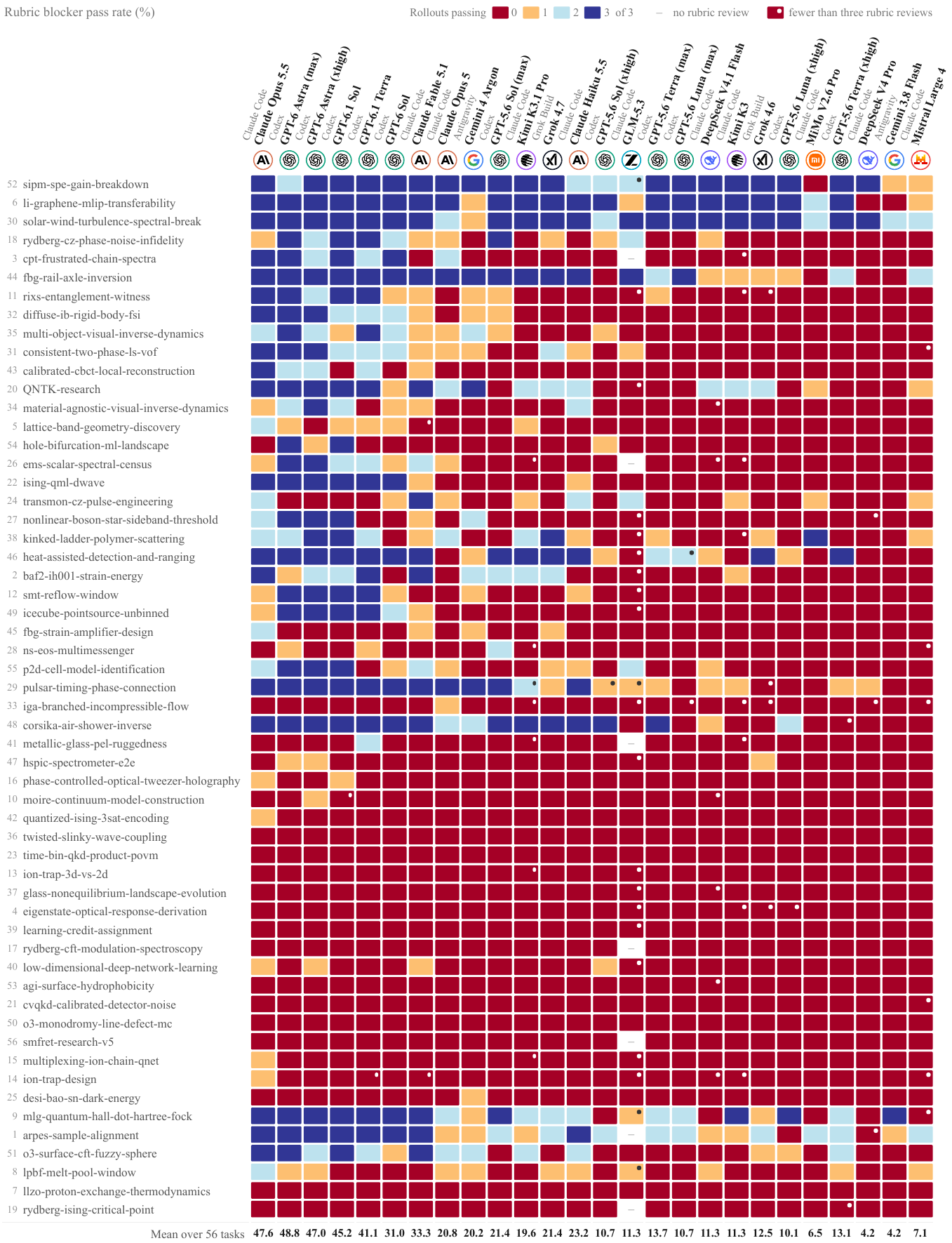}
\caption{Rubric blocker pass rate per task and model. The bottom row is each model's mean over all \FPTaskCount{} tasks. Unavailable outcomes contribute zero to the aggregate.}
\label{fig:per-task-blockers}
\end{figure}

\begin{figure}[p]
\centering
\includegraphics[width=\linewidth]{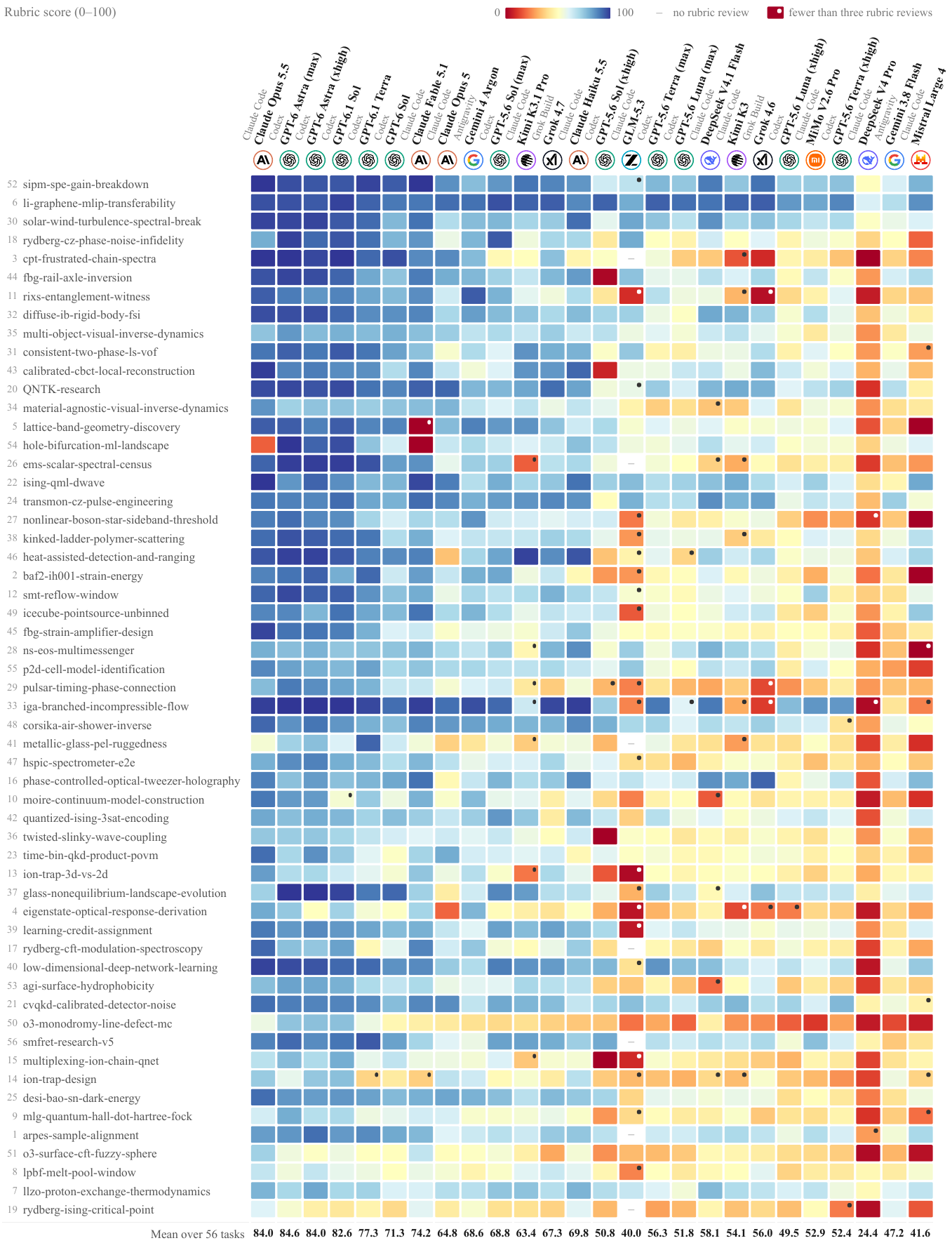}
\caption{Rubric score per task and model. The weighted non-blocker score (Equation \ref{eq:score}) before the verifier and blocker gates, averaged over three rollouts. The bottom row is each model's mean over all \FPTaskCount{} tasks.}
\label{fig:per-task-rubric}
\end{figure}

\begin{figure}[p]
\centering
\includegraphics[width=\linewidth]{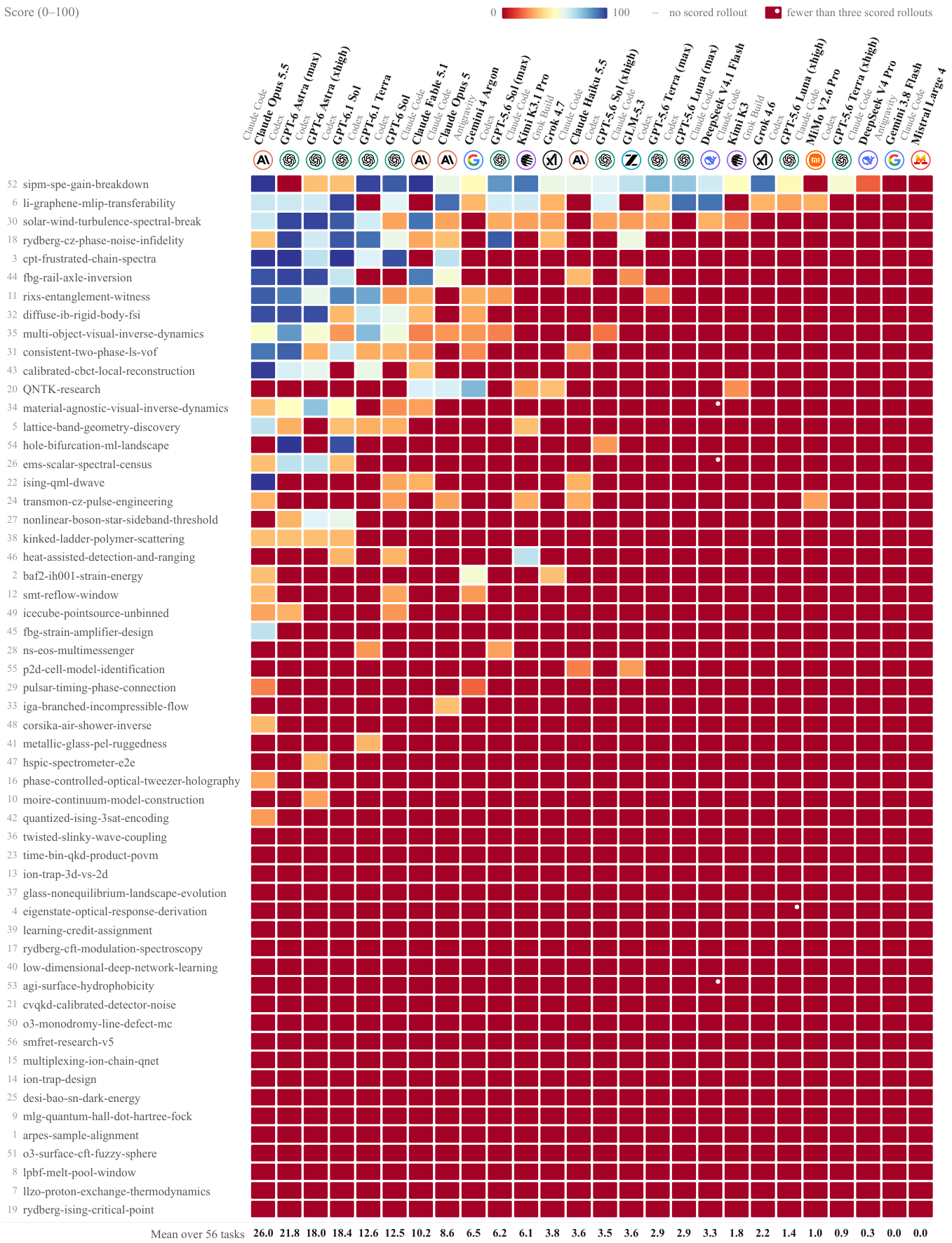}
\caption{Score  (Equation \ref{eq:score}) per task and model. The bottom row reproduces Score in Table~\ref{tab:main-results}. Rows follow Figure~\ref{fig:per-task-verifier}. Missing rollouts count as zero.}
\label{fig:per-task-score}
\end{figure}

Figures~\ref{fig:per-task-verifier}--\ref{fig:per-task-score} break the results of Table~\ref{tab:main-results} down by task for all \FPModelCount{} model configurations. Each cell summarizes the three designated rollouts of one model on one task. As in Table~\ref{tab:main-results}, an unavailable rollout counts as zero. The bottom row of each figure is the model's mean over all \FPTaskCount{} tasks. The four following figures show:

\begin{itemize}
\item \textbf{Verifier pass rate} (Figure~\ref{fig:per-task-verifier}): the percentage of rollouts in which every applicable deterministic test passes. Its bottom row is the deterministic Pass@1 of Table~\ref{tab:main-results}.
\item \textbf{Rubric blocker pass rate} (Figure~\ref{fig:per-task-blockers}): the percentage of rollouts in which every rubric blocker passes.
\item \textbf{Rubric score} (Figure~\ref{fig:per-task-rubric}): the weighted non-blocker score in Equation \ref{eq:score}, before the verifier and blocker gates are applied.
\item \textbf{Score} (Figure~\ref{fig:per-task-score}): the rubric score (Equation \ref{eq:score}) of rollouts that pass the verifier and every blocker, and zero otherwise. Its bottom row is the Score of Table~\ref{tab:main-results}.
\end{itemize}

 \clearpage
\section{Failure-Mode Analysis}
\label{app:failure-labeling}

The following table shows a detailed explanation of the failure mode taxonomy that was extracted on the latest GPT-6 Astra (\texttt{xhigh}, \texttt{codex-acp}) cohort based on the method described in Section \ref{sec:reubric_reliability}.

\begingroup
\definecolor{fmPHY}{HTML}{8D4149}
\definecolor{fmLIT}{HTML}{68527C}
\definecolor{fmINS}{HTML}{355D82}
\definecolor{fmREAL}{HTML}{7D6038}
\definecolor{fmINT}{HTML}{355F57}
\definecolor{fmCOH}{HTML}{84546B}
\definecolor{fmRIG}{HTML}{886449}
\definecolor{fmFALSE}{HTML}{B14545}
\definecolor{fmMIS}{HTML}{777777}
\definecolor{fmSP}{HTML}{313695}
\definecolor{fmRRMIS}{HTML}{C1694B}
\definecolor{fmEXE}{HTML}{526373}
\definecolor{fmHAL}{HTML}{875500}
\definecolor{fmHACK}{HTML}{623C83}
\definecolor{fmPass}{HTML}{3E6B54}
\definecolor{fmNA}{HTML}{60686E}
\newcommand{\fmTag}[2]{\textcolor{#1}{\textbf{#2}}}
\newcommand{\fmSecondary}[2]{\textcolor{#1}{#2}}
\begingroup
\small
\hyphenpenalty=10000\exhyphenpenalty=0
\setlength{\tabcolsep}{3pt}
\renewcommand{\arraystretch}{1.08}
\setlength{\LTleft}{0pt}\setlength{\LTright}{0pt}
\setlength{\LTcapwidth}{\textwidth}
\newcommand{\fmDocBullet}[1]{\begingroup\hangindent=0.9em\hangafter=1\noindent\makebox[0.9em][l]{\textbullet}#1\par\endgroup}
\begin{longtable}{@{}>{\raggedright\arraybackslash}p{0.20\textwidth}>{\raggedright\arraybackslash}p{0.09\textwidth}>{\raggedright\arraybackslash}p{\dimexpr0.49\textwidth-6\tabcolsep\relax}>{\raggedright\arraybackslash}p{0.22\textwidth}@{}}
\caption{Failure mode taxonomy.}\label{tab:failure-taxonomy}\\
\toprule
\textbf{Tag} & \textbf{Label} & \textbf{Description} & \textbf{Examples} \\
\midrule
\endfirsthead
\caption[]{Failure mode taxonomy (continued).}\\
\toprule
\textbf{Tag} & \textbf{Label} & \textbf{Description} & \textbf{Examples} \\
\midrule
\endhead
\midrule\multicolumn{4}{r}{\small Continued on next page}\\
\endfoot
\bottomrule
\endlastfoot
\rowcolor{altrow}
Get the physics wrong & \fmTag{fmPHY}{PHY} & \fmDocBullet{Subtle mistakes in the formulas or theory applied;} \fmDocBullet{Agents do not have such physics knowledge (no signal when checking agents' trajectories).} & \texttt{mlg-\allowbreak quantum-\allowbreak hall-\allowbreak dot-\allowbreak hartree-\allowbreak fock}\par\smallskip \texttt{material-\allowbreak agnostic-\allowbreak visual-\allowbreak inverse-\allowbreak dynamics} \\
Scientific presenting & \fmTag{fmSP}{SP} & Agents have a good understanding of the task, but the paper writing is problematic and cannot present the research professionally.\par & \texttt{rixs-\allowbreak entanglement-\allowbreak witness} \\
\rowcolor{altrow}
False grading & \fmTag{fmRRMIS}{RRMIS} & The reviewer agent's rubric grading does not align with human researchers' grading. For example, the rubric is correct and the worker agent does the right thing, but the reviewer agent makes a mistake (often by not following the instructions properly).\par & \texttt{li-\allowbreak graphene-\allowbreak mlip-\allowbreak transferability} \\
Literature mistake & \fmTag{fmLIT}{LIT} & \fmDocBullet{Agents cannot find the important papers (specified by the task author in rubrics as important papers that should be found) in the research stage;} \fmDocBullet{In \texttt{paper.pdf}, agents cannot use the literature to support their claims correctly.} & \texttt{arpes-\allowbreak sample-\allowbreak alignment}\par\smallskip \texttt{calibrated-\allowbreak cbct-\allowbreak local-\allowbreak reconstruction} \\
\rowcolor{altrow}
Agents do not follow \texttt{task.md} properly & \fmTag{fmINS}{INS} & \fmDocBullet{Agents miss certain instructions in \texttt{task.md} (do not make the judgement based only on agents' artifacts; check their trajectories to see if agents ever recognize the instructions).} \fmDocBullet{Agents don't understand or misunderstand the instructions.} \par\vspace{0.35\baselineskip} In the example task, agents do not get that ``we work in a nearest-neighbor-Rydberg-blockade regime'' implies they should make approximations to the Hamiltonian or perform postselection.\par & \texttt{glass-\allowbreak nonequilibrium-\allowbreak landscape-\allowbreak evolution}\par\smallskip \texttt{low-\allowbreak dimensional-\allowbreak deep-\allowbreak network-\allowbreak learning}\par\smallskip \texttt{rydberg-\allowbreak ising-\allowbreak critical-\allowbreak point}\par{\normalfont\scriptsize (Trials 1--3; supporting; task-dominant: REAL)} \\
Agents do not understand the reality constraints\par & \fmTag{fmREAL}{REAL} & In the example task, agents do not get the reality constraints of the simulated device.\par & \texttt{rydberg-\allowbreak ising-\allowbreak critical-\allowbreak point} \\
\rowcolor{altrow}
Missing physics expert intuition or experience & \fmTag{fmINT}{INT} & In the example task rollouts, agents do not consider Rydberg decay when analyzing sources of error. An AMO physicist would naturally consider this mechanism, suggesting a gap in the agents' physical intuition.\par \par\vspace{0.35\baselineskip} Also, sometimes in the thinking traces, agents are aware of the instructions but still fail in the end since they lack the expert-level physics intuition needed to apply them correctly.\par & \texttt{multiplexing-\allowbreak ion-\allowbreak chain-\allowbreak qnet}\par\smallskip \texttt{lpbf-\allowbreak melt-\allowbreak pool-\allowbreak window}\par\smallskip \texttt{hspic-\allowbreak spectrometer-\allowbreak e2e}\par\smallskip \texttt{rydberg-\allowbreak cft-\allowbreak modulation-\allowbreak spectroscopy}\par{\normalfont\scriptsize (Trial 1; supporting; task-dominant: REAL)} \\
Incoherent reasoning & \fmTag{fmCOH}{COH} & \fmDocBullet{Contradictory logic chains or arguments.} \par\vspace{0.35\baselineskip} In the example task rollouts, agents realize that they are wrong, but they later forget this.\par & \texttt{kinked-\allowbreak ladder-\allowbreak polymer-\allowbreak scattering} \\
\rowcolor{altrow}
Lack of rigorous thinking & \fmTag{fmRIG}{RIG} & \fmDocBullet{Circular self-verification.} \fmDocBullet{Agents do not consider enough edge cases, so the program or method only works for limited cases.} \par\vspace{0.35\baselineskip} In the example task rollouts, agents' self-verification method depends on their own assumption, which is wrong, so they can never find their mistakes.\par & \texttt{p2d-\allowbreak cell-\allowbreak model-\allowbreak identification}\par\smallskip \texttt{desi-\allowbreak bao-\allowbreak sn-\allowbreak dark-\allowbreak energy}\par\smallskip \texttt{time-\allowbreak bin-\allowbreak qkd-\allowbreak product-\allowbreak povm}\par\smallskip \texttt{hole-\allowbreak bifurcation-\allowbreak ml-\allowbreak landscape} \\
False negative & \fmTag{fmFALSE}{FALSE} & Agents do the task correctly but still fail because of a task-design issue (unclear \texttt{task.md}, a verifier or rubric that contradicts the instructions, an overly strict tolerance).\par & \texttt{phase-\allowbreak controlled-\allowbreak optical-\allowbreak tweezer-\allowbreak holography} \\
\rowcolor{altrow}
Misgrade & \fmTag{fmMIS}{MIS} & Certain test or rubric results are not correct for this trial, but they do not influence whether this trial passes or not.\par & \texttt{arpes-\allowbreak sample-\allowbreak alignment}\par{\normalfont\scriptsize (Trials 1,3; task-dominant: LIT)} \\
Agents' execution mistakes & \fmTag{fmEXE}{EXE} & Agents cannot write proper code or use the tools properly, etc.\par & \texttt{icecube-\allowbreak pointsource-\allowbreak unbinned} \\
\rowcolor{altrow}
Hallucination & \fmTag{fmHAL}{HAL} & Agents fabricate evidence, data, experiments, literature, etc.\par & \texttt{nonlinear-\allowbreak boson-\allowbreak star-\allowbreak sideband-\allowbreak threshold}\par{\normalfont\scriptsize (Trial 1; task-dominant: RIG)} \\
Reward hacking & \fmTag{fmHACK}{HACK} & Agents try to hack to get the full reward by:\par \fmDocBullet{Trying to change verifiers or hack rubrics to find answers;} \fmDocBullet{Trying to directly find answers online and copy them to the final workspace to pretend the job has been done, etc.} & - \\
\end{longtable}
\endgroup
 \subsection{Final Trial Labels and Main Findings}
Each trial retains all labels and their order from the final review sheet, including locally rerun trials. PASS may coexist with other annotations. $^{\mathrm{V}}$ preserves a verifier marker from the sheet. A dash in the finding column means no finding was supplied. For \texttt{baf2-ih001-strain-energy}, the three columns correspond to local Trials 4--6. Task numbers follow Table~\ref{tab:task-domains}.

\definecolor{fmOutcomeChange}{RGB}{0,255,255}
\small
\setlength{\tabcolsep}{2pt}
\setlength{\LTleft}{0pt}\setlength{\LTright}{0pt}
\setlength{\LTcapwidth}{\textwidth}
\renewcommand{\arraystretch}{1.08}
\begin{longtable}{@{}>{\raggedleft\arraybackslash}p{0.03\textwidth}>{\raggedright\arraybackslash}p{0.30\textwidth}>{\centering\arraybackslash}p{0.09\textwidth}>{\centering\arraybackslash}p{0.09\textwidth}>{\centering\arraybackslash}p{0.09\textwidth}>{\raggedright\arraybackslash}p{\dimexpr0.40\textwidth-10\tabcolsep\relax}@{}}
\caption{Final review-sheet labels for 56 tasks. Cyan cells (eight FALSE, one RRMIS) identify the 9 trials whose final pass/fail outcomes should change; MIS denotes verifier or rubric errors that do not change the final outcome.}\phantomsection\label{tab:failure-tasks}\\
\toprule
\textbf{\#} & \textbf{Task} & \textbf{Trial 1} & \textbf{Trial 2} & \textbf{Trial 3} & \textbf{Main finding} \\
\midrule
\endfirsthead
\caption[]{Final review-sheet labels and task findings (continued).}\\
\toprule
\textbf{\#} & \textbf{Task} & \textbf{Trial 1} & \textbf{Trial 2} & \textbf{Trial 3} & \textbf{Main finding} \\
\midrule
\endhead
\midrule\multicolumn{6}{r}{\small Continued on next page}\\
\endfoot
\bottomrule
\endlastfoot
\addlinespace[4pt]
\multicolumn{6}{@{}l}{\textbf{Condensed Matter Physics}}\\*
\addlinespace[2pt]
1 & \texttt{arpes-\allowbreak sample-\allowbreak alignment} & \fmTag{fmRIG}{RIG}\newline \fmTag{fmLIT}{LIT}\newline \fmTag{fmMIS}{MIS} & \fmTag{fmLIT}{LIT}\newline \fmTag{fmRIG}{RIG} & \fmTag{fmLIT}{LIT}\newline \fmTag{fmMIS}{MIS} & Agents recognized tilt--translation coupling but missed pose-recovery tolerances and omitted electron-optics or inverse-optimization references. Reviewers overlooked executed one-coordinate scans in Trials 1 and 3, although both still failed pose recovery. \\
\rowcolor{altrow}
2 & \texttt{baf2-\allowbreak ih001-\allowbreak strain-\allowbreak energy} & \fmTag{fmINT}{INT}\newline \fmTag{fmRIG}{RIG} & \fmTag{fmINT}{INT}\newline \fmTag{fmRIG}{RIG} & \fmTag{fmRIG}{RIG} & Displayed Trials 1--2 insufficiently justified bulk-like ice selection and failed AgI transfer. Displayed Trial 3 passed AgI but did not justify its selected ice population despite crop-dependent results. \\
3 & \texttt{cpt-\allowbreak frustrated-\allowbreak chain-\allowbreak spectra} & \fmTag{fmPass}{PASS} & \fmTag{fmPass}{PASS} & \fmTag{fmPass}{PASS} & - \\
\rowcolor{altrow}
4 & \texttt{eigenstate-\allowbreak optical-\allowbreak response-\allowbreak derivation} & \fmTag{fmINS}{INS}\newline \fmTag{fmMIS}{MIS} & \fmTag{fmINS}{INS}\newline \fmTag{fmMIS}{MIS} & \fmTag{fmINS}{INS} & Agents did not start from the required current-density operator. PDF extraction and Lean environment setup also produced erroneous checks, while other verifier and blocker failures remained. \\
5 & \texttt{lattice-\allowbreak band-\allowbreak geometry-\allowbreak discovery} & \fmTag{fmRIG}{RIG}\newline \fmTag{fmLIT}{LIT} & \cellcolor{fmOutcomeChange}\fmTag{fmFALSE}{FALSE} & \fmTag{fmINS}{INS}\newline \fmTag{fmRIG}{RIG} & Trial 1 proved the scaling limit, but an early mistake spoiled its numerics; Trial 2's correct numerics hit a run-time error because the task requested too large a grid, a false negative, and Trial 3 met the same problem. \\
\rowcolor{altrow}
6 & \texttt{li-\allowbreak graphene-\allowbreak mlip-\allowbreak transferability} & \fmTag{fmPass}{PASS}\newline \fmTag{fmRIG}{RIG} & \cellcolor{fmOutcomeChange}\fmTag{fmRRMIS}{RRMIS}\newline \fmTag{fmRIG}{RIG} & \fmTag{fmPass}{PASS} & Trials 1 and 2 computed frame errors correctly but did not discuss why two of the 25 frames had larger errors; Trial 2's verifier missed a correctly printed adsorption energy after OCR dropped its minus sign. Trial 3's paper needed better writing. \\
7 & \texttt{llzo-\allowbreak proton-\allowbreak exchange-\allowbreak thermodynamics} & \fmTag{fmRIG}{RIG}\newline \fmTag{fmLIT}{LIT} & \fmTag{fmRIG}{RIG}\newline \fmTag{fmLIT}{LIT} & \fmTag{fmRIG}{RIG}\newline \fmTag{fmLIT}{LIT} & None of the agents reported the trigonal planar Li that belong to neither cavity, and two built their parent lattice on coordinates with no traceable source. \\
\rowcolor{altrow}
8 & \texttt{lpbf-\allowbreak melt-\allowbreak pool-\allowbreak window} & \fmTag{fmRIG}{RIG}\newline \fmTag{fmINT}{INT} & \fmTag{fmINT}{INT} & \fmTag{fmINT}{INT} & Agents recommended unreliable process windows, retaining unstable candidates and marking unsafe settings as stable\_screening; Trial 2 nearly passed but claimed an unstable setting stable. \\
9 & \texttt{mlg-\allowbreak quantum-\allowbreak hall-\allowbreak dot-\allowbreak hartree-\allowbreak fock} & \fmTag{fmPHY}{PHY}\newline \fmTag{fmRIG}{RIG} & \fmTag{fmPHY}{PHY}\newline \fmTag{fmRIG}{RIG} & \fmTag{fmPHY}{PHY}\newline \fmTag{fmRIG}{RIG} & Agents left the external confinement unscreened, omitting the RPA screening that this simulation setting required, among other physics mistakes; an inexperienced human might also have made this omission. \\
\rowcolor{altrow}
10 & \texttt{moire-\allowbreak continuum-\allowbreak model-\allowbreak construction} & \fmTag{fmINS}{INS}\newline \fmTag{fmMIS}{MIS} & \fmTag{fmPass}{PASS}\newline \fmTag{fmINS}{INS} & \fmTag{fmINS}{INS}\newline \fmTag{fmMIS}{MIS} & Agents passed every numerical test but mostly reached the answer through a special symmetry argument, too specialized for this model class, although the task asked for a universal low-energy reduction instead. \\
11 & \texttt{rixs-\allowbreak entanglement-\allowbreak witness} & \fmTag{fmSP}{SP}\newline \fmTag{fmINT}{INT} & \fmTag{fmPass}{PASS}\newline \fmTag{fmSP}{SP} & \fmTag{fmPass}{PASS}\newline \fmTag{fmSP}{SP}\newline \fmTag{fmINT}{INT} & Agents got the physics right but left out important assumptions a publishable paper should discuss; in Trials 1 and 3 they also simulated correctly but missed the phase transition in their results. \\
\rowcolor{altrow}
12 & \texttt{smt-\allowbreak reflow-\allowbreak window} & \fmTag{fmINT}{INT}\newline \fmTag{fmREAL}{REAL} & \fmTag{fmINT}{INT}\newline \fmTag{fmREAL}{REAL} & \fmTag{fmINT}{INT}\newline \fmTag{fmREAL}{REAL} & Agents missed real-world constraints and lacked expert intuition (e.g., that excess solder raises the risk of shorts); their bridge and cold-joint scores did not respond to clearly risky paste or thermal conditions, and Trial 2 started from random guesses. \\
\addlinespace[4pt]
\multicolumn{6}{@{}l}{\textbf{AMO Physics}}\\*
\addlinespace[2pt]
13 & \texttt{ion-\allowbreak trap-\allowbreak 3d-\allowbreak vs-\allowbreak 2d} & \fmTag{fmINT}{INT}\newline \fmTag{fmMIS}{MIS} & \fmTag{fmINT}{INT} & \fmTag{fmINT}{INT} & Agents solved the electrostatics correctly but failed to connect static trap depth to experimental ion loss. Trial 1 also received inconsistent grading on pseudopotential validity. \\
\rowcolor{altrow}
14 & \texttt{ion-\allowbreak trap-\allowbreak design} & \cellcolor{fmOutcomeChange}\fmTag{fmFALSE}{FALSE} & \fmTag{fmRIG}{RIG}\newline \fmTag{fmCOH}{COH} & \fmTag{fmINT}{INT}\newline \fmTag{fmRIG}{RIG}\newline \fmTag{fmCOH}{COH} & Trial 1 met trap-design and operating-configuration requirements; Trial 2 met only the former. Trial 3 applied inappropriate gapless electrostatics to a slitted trap and lacked independent verification despite recognizing its importance. \\
15 & \texttt{multiplexing-\allowbreak ion-\allowbreak chain-\allowbreak qnet} & \fmTag{fmINT}{INT} & \fmTag{fmINT}{INT} & \fmTag{fmINT}{INT} & Agents could not fit Rabi-flop data with a model combining coherent excitation and thermal heating. \\
\rowcolor{altrow}
16 & \texttt{phase-\allowbreak controlled-\allowbreak optical-\allowbreak tweezer-\allowbreak holography} & \cellcolor{fmOutcomeChange}\fmTag{fmFALSE}{FALSE} & \cellcolor{fmOutcomeChange}\fmTag{fmFALSE}{FALSE} & \cellcolor{fmOutcomeChange}\fmTag{fmFALSE}{FALSE} & - \\
17 & \texttt{rydberg-\allowbreak cft-\allowbreak modulation-\allowbreak spectroscopy} & \fmTag{fmINT}{INT}\newline \fmTag{fmREAL}{REAL}\newline \fmTag{fmPHY}{PHY} & \fmTag{fmREAL}{REAL} & \fmTag{fmREAL}{REAL} & All trials omitted device calibration. Trial 1 additionally misunderstood CFT, inadequately assessed Rydberg blockade, and omitted decay as an error source. Its rubric review also contained both false-negative and false-positive judgments. \\
\rowcolor{altrow}
18 & \texttt{rydberg-\allowbreak cz-\allowbreak phase-\allowbreak noise-\allowbreak infidelity} & \fmTag{fmCOH}{COH}\newline \fmTag{fmINT}{INT}\newline \fmTag{fmRIG}{RIG}\newline \fmTag{fmEXE}{EXE}\newline (\fmTag{fmFALSE}{FALSE}) & \fmTag{fmPass}{PASS}\newline \fmTag{fmRIG}{RIG}\newline \fmTag{fmFALSE}{FALSE} & \fmTag{fmPass}{PASS}\newline \fmTag{fmRIG}{RIG}\newline \fmTag{fmINT}{INT} & All trials passed the verifier tests. Trial 1 failed a blocker by justifying its answer only with in-loop noise squashing, which its own data contradicted; Trials 1 and 3 did not state that the PDH servo bump causes the infidelity gap. \\
19 & \texttt{rydberg-\allowbreak ising-\allowbreak critical-\allowbreak point} & \fmTag{fmREAL}{REAL}\newline \fmTag{fmMIS}{MIS}\newline \fmTag{fmINS}{INS}\newline \fmTag{fmLIT}{LIT} & \fmTag{fmREAL}{REAL}\newline \fmTag{fmMIS}{MIS}\newline \fmTag{fmINS}{INS}\newline \fmTag{fmLIT}{LIT} & \fmTag{fmREAL}{REAL}\newline \fmTag{fmMIS}{MIS}\newline \fmTag{fmINS}{INS}\newline \fmTag{fmLIT}{LIT} & Agents exceeded realistic shot budgets, omitted calibration, misunderstood open-boundary finite-size numerics, and weakly synthesized literature. Grading also relied on rigid section-word checks and an unclear nearest-neighbor blockade assumption. \\
\addlinespace[4pt]
\multicolumn{6}{@{}l}{\textbf{Quantum Information}}\\*
\addlinespace[2pt]
\rowcolor{altrow}
20 & \texttt{QNTK-\allowbreak research} & \fmTag{fmMIS}{MIS}\newline \fmTag{fmRIG}{RIG}\newline \fmTag{fmINS}{INS} & \fmTag{fmMIS}{MIS}\newline \fmTag{fmRIG}{RIG}\newline \fmTag{fmINS}{INS} & \fmTag{fmINS}{INS}\newline \fmTag{fmMIS}{MIS}\newline \fmTag{fmRIG}{RIG} & Agents did not use the required circuit model. Their self-verification reused their own assumptions and provided weak checks. \\
21 & \texttt{cvqkd-\allowbreak calibrated-\allowbreak detector-\allowbreak noise} & \fmTag{fmPHY}{PHY}\newline \fmTag{fmLIT}{LIT}\newline \fmTag{fmRIG}{RIG} & \fmTag{fmPHY}{PHY}\newline \fmTag{fmLIT}{LIT}\newline \fmTag{fmRIG}{RIG} & \fmTag{fmPHY}{PHY}\newline \fmTag{fmLIT}{LIT}\newline \fmTag{fmRIG}{RIG} & Agents missed the intended detector model, using security assumptions that differed from the reference and giving a wrong modulation variance and key rates negative beyond 100 km; key papers appeared late, not in the introduction, and the PLOB bound went unchecked. \\
\rowcolor{altrow}
22 & \texttt{ising-\allowbreak qml-\allowbreak dwave} & \fmTag{fmPass}{PASS} & \fmTag{fmPass}{PASS}\newline \fmTag{fmLIT}{LIT} & \fmTag{fmPass}{PASS} & The physics and optimization were correct in all trials; failures concerned the analysis and write-up. \\
23 & \texttt{time-\allowbreak bin-\allowbreak qkd-\allowbreak product-\allowbreak povm} & \fmTag{fmRIG}{RIG}\newline \fmTag{fmLIT}{LIT} & \fmTag{fmRIG}{RIG}\newline \fmTag{fmINS}{INS}\newline \fmTag{fmLIT}{LIT} & \fmTag{fmRIG}{RIG}\newline \fmTag{fmLIT}{LIT} & Agents computed the numerical results correctly, but their papers omitted the security implications and why Lavie--Lim's extra-vacuum-pulse method is needed, and missed key prior work; Trial 2 also skipped the IID discussion and once tried to access papers past the literature cutoff. \\
\rowcolor{altrow}
24 & \texttt{transmon-\allowbreak cz-\allowbreak pulse-\allowbreak engineering} & \fmTag{fmINT}{INT}\newline \fmTag{fmCOH}{COH} & \fmTag{fmINT}{INT}\newline \fmTag{fmLIT}{LIT}\newline \fmTag{fmCOH}{COH} & \fmTag{fmINT}{INT}\newline \fmTag{fmCOH}{COH} & Agents implemented the physics correctly but failed line-accuracy tests: they recognized noise in the final third of their results yet published it unsmoothed. Trial 2 used a method without properly citing its paper. \\
\addlinespace[4pt]
\multicolumn{6}{@{}l}{\textbf{Astrophysics}}\\*
\addlinespace[2pt]
25 & \texttt{desi-\allowbreak bao-\allowbreak sn-\allowbreak dark-\allowbreak energy} & \fmTag{fmRIG}{RIG}\newline \fmTag{fmINT}{INT}\newline \fmTag{fmINS}{INS} & \fmTag{fmRIG}{RIG}\newline \fmTag{fmINT}{INT}\newline \fmTag{fmINS}{INS} & \fmTag{fmRIG}{RIG}\newline \fmTag{fmINT}{INT}\newline \fmTag{fmINS}{INS} & The pipeline always returns a precise BAO scale, even for spectra with no acoustic feature, and its weak-signal intervals are uncalibrated; closure is run but not reported in the required form. Downstream cosmology, SN selection, degeneracies and null calibration are correct in all three trials. \\
\rowcolor{altrow}
26 & \texttt{ems-\allowbreak scalar-\allowbreak spectral-\allowbreak census} & \fmTag{fmRIG}{RIG}\newline \fmTag{fmPass}{PASS} & \fmTag{fmPass}{PASS} & \fmTag{fmPass}{PASS} & Trial 1 missed hidden edge cases. Trial 2 explicitly considered the edge case; Trial 3 avoided it through a different design. \\
27 & \texttt{nonlinear-\allowbreak boson-\allowbreak star-\allowbreak sideband-\allowbreak threshold} & \fmTag{fmHAL}{HAL}\newline \fmTag{fmINT}{INT}\newline \fmTag{fmRIG}{RIG}\newline \fmTag{fmPHY}{PHY}\newline \fmTag{fmSP}{SP}\newline \fmTag{fmLIT}{LIT} & \fmTag{fmRIG}{RIG}\newline \fmTag{fmINT}{INT}\newline \fmTag{fmSP}{SP}\newline \fmTag{fmLIT}{LIT} & \fmTag{fmPass}{PASS} & Trial 1's mode classifier missed edge cases and miscounted nodes. Its grading also depended on a precision requirement absent from task.md. \\
\rowcolor{altrow}
28 & \texttt{ns-\allowbreak eos-\allowbreak multimessenger} & \cellcolor{fmOutcomeChange}\fmTag{fmFALSE}{FALSE} & \fmTag{fmRIG}{RIG} & \fmTag{fmPass}{PASS} & Trial 1 was penalized for an ambiguous prior convention. Trial 2 inconsistently enforced stable-branch selection, and its input guard rejected hidden data. The physics was largely correct; papers omitted equations and caveats present in code. \\
29 & \texttt{pulsar-\allowbreak timing-\allowbreak phase-\allowbreak connection} & \fmTag{fmRIG}{RIG}\newline \fmTag{fmMIS}{MIS} & \fmTag{fmRIG}{RIG}\newline \fmTag{fmMIS}{MIS} & \fmTag{fmRIG}{RIG}\newline \fmTag{fmMIS}{MIS} & Agents applied a standard threshold without adapting or validating it for this specific case. \\
\rowcolor{altrow}
30 & \texttt{solar-\allowbreak wind-\allowbreak turbulence-\allowbreak spectral-\allowbreak break} & \fmTag{fmPass}{PASS} & \fmTag{fmPass}{PASS} & \fmTag{fmPass}{PASS} & All three trials passed. \\
\addlinespace[4pt]
\multicolumn{6}{@{}l}{\textbf{Fluid Dynamics}}\\*
\addlinespace[2pt]
31 & \texttt{consistent-\allowbreak two-\allowbreak phase-\allowbreak ls-\allowbreak vof} & \fmTag{fmFALSE}{FALSE} & \cellcolor{fmOutcomeChange}\fmTag{fmFALSE}{FALSE} & \cellcolor{fmOutcomeChange}\fmTag{fmFALSE}{FALSE} & Trial 2's verifier failure may stem from an unspecified accuracy requirement that allowed premature stopping. \\
\rowcolor{altrow}
32 & \texttt{diffuse-\allowbreak ib-\allowbreak rigid-\allowbreak body-\allowbreak fsi} & \fmTag{fmPass}{PASS}\newline \fmTag{fmINT}{INT}\newline \fmTag{fmMIS}{MIS} & \fmTag{fmPass}{PASS} & \fmTag{fmPass}{PASS} & Agents implemented surface versus volumetric forcing as superficial point-placement differences rather than using the correct fractional-step stage. Reviewers also penalized coding before a detailed plan, which the task did not require. \\
33 & \texttt{iga-\allowbreak branched-\allowbreak incompressible-\allowbreak flow} & \fmTag{fmINT}{INT}\newline (\fmTag{fmREAL}{REAL})\newline (\fmTag{fmRIG}{RIG}) & \fmTag{fmINT}{INT}\newline (\fmTag{fmREAL}{REAL})\newline (\fmTag{fmRIG}{RIG}) & \fmTag{fmINT}{INT}\newline (\fmTag{fmREAL}{REAL})\newline (\fmTag{fmRIG}{RIG}) & Agents failed hidden cases with different geometries: they lacked a global understanding of the final model, whose issues a global overview in professional software would reveal, and gave edge cases too little thought. \\
\rowcolor{altrow}
34 & \texttt{material-\allowbreak agnostic-\allowbreak visual-\allowbreak inverse-\allowbreak dynamics} & \fmTag{fmPHY}{PHY} & \fmTag{fmPHY}{PHY} & \fmTag{fmPHY}{PHY} & Agents failed to design sufficient supervision for stable constitutive learning. \\
35 & \texttt{multi-\allowbreak object-\allowbreak visual-\allowbreak inverse-\allowbreak dynamics} & \fmTag{fmPHY}{PHY}\newline \fmTag{fmRIG}{RIG}\newline \fmTag{fmINS}{INS} & \fmTag{fmPHY}{PHY}\newline \fmTag{fmRIG}{RIG}\newline \fmTag{fmINS}{INS} & \fmTag{fmPHY}{PHY}\newline \fmTag{fmRIG}{RIG}\newline \fmTag{fmINS}{INS} & Agents failed to distinguish constitutive errors from geometry and contact errors. \\
\rowcolor{altrow}
36 & \texttt{twisted-\allowbreak slinky-\allowbreak wave-\allowbreak coupling} & \fmTag{fmINT}{INT}\newline \fmTag{fmLIT}{LIT} & \fmTag{fmINT}{INT}\newline \fmTag{fmLIT}{LIT} & \fmTag{fmINT}{INT}\newline \fmTag{fmLIT}{LIT} & Agents treated the spring as a whole instead of analyzing transverse motion at the cross-sectional and coil scales. They therefore failed to explain transverse--torsional coupling and provided only qualitative relations and overall trend fits. \\
\addlinespace[4pt]
\multicolumn{6}{@{}l}{\textbf{Statistical Physics}}\\*
\addlinespace[2pt]
37 & \texttt{glass-\allowbreak nonequilibrium-\allowbreak landscape-\allowbreak evolution} & \fmTag{fmINS}{INS} & \fmTag{fmINS}{INS} & \fmTag{fmINS}{INS} & Agents passed all verifier tests but omitted the required comparison of factorized and dependence-preserving kinetic models and its implications for the ageing--rejuvenation crossover; Trial 2 also left activation--relaxation dependence unassessed. \\
\rowcolor{altrow}
38 & \texttt{kinked-\allowbreak ladder-\allowbreak polymer-\allowbreak scattering} & \fmTag{fmINT}{INT}\newline \fmTag{fmCOH}{COH}\newline \fmTag{fmRIG}{RIG} & \fmTag{fmINT}{INT}\newline \fmTag{fmCOH}{COH}\newline \fmTag{fmRIG}{RIG} & \fmTag{fmINT}{INT}\newline \fmTag{fmCOH}{COH}\newline \fmTag{fmRIG}{RIG} & Agents retained an incorrect sign rule while tuning other parameters, then claimed validation despite documenting the unresolved problem in the appendix. \\
39 & \texttt{learning-\allowbreak credit-\allowbreak assignment} & \fmTag{fmINS}{INS} & \fmTag{fmINS}{INS} & \fmTag{fmINS}{INS} & Agents gave results around the requested limit rather than under it and, when told to apply their model to other problems, used another method. \\
\rowcolor{altrow}
40 & \texttt{low-\allowbreak dimensional-\allowbreak deep-\allowbreak network-\allowbreak learning} & \fmTag{fmINS}{INS}\newline \fmTag{fmLIT}{LIT} & \fmTag{fmINS}{INS}\newline \fmTag{fmINT}{INT}\newline \fmTag{fmLIT}{LIT} & \fmTag{fmINS}{INS}\newline \fmTag{fmPHY}{PHY}\newline \fmTag{fmINT}{INT}\newline \fmTag{fmLIT}{LIT} & Trial 1 chose a wrong algorithm by ignoring conditions in task.md. Agents performed the mathematics correctly but selected the less-efficient, unshared design and did not recognize that their theory required a simplifying assumption. \\
41 & \texttt{metallic-\allowbreak glass-\allowbreak pel-\allowbreak ruggedness} & \fmTag{fmINS}{INS}\newline \fmTag{fmRIG}{RIG} & \fmTag{fmINS}{INS}\newline \fmTag{fmRIG}{RIG} & \fmTag{fmPass}{PASS} & Agents recognized the described differences but did not identify a quantitative measure with physical implications; one failed trial also omitted the required raw energy and force measurements. \\
\rowcolor{altrow}
42 & \texttt{quantized-\allowbreak ising-\allowbreak 3sat-\allowbreak encoding} & \fmTag{fmPass}{PASS}\newline \fmTag{fmINS}{INS}\newline \fmTag{fmLIT}{LIT} & \fmTag{fmPass}{PASS}\newline \fmTag{fmINS}{INS}\newline \fmTag{fmLIT}{LIT} & \fmTag{fmPass}{PASS}\newline \fmTag{fmINS}{INS} & All three trials solved the formulas; the failures concerned the papers. \\
\addlinespace[4pt]
\multicolumn{6}{@{}l}{\textbf{Photonics \& Imaging}}\\*
\addlinespace[2pt]
43 & \texttt{calibrated-\allowbreak cbct-\allowbreak local-\allowbreak reconstruction} & \fmTag{fmINS}{INS}\newline \fmTag{fmLIT}{LIT} & \fmTag{fmINS}{INS}\newline \fmTag{fmLIT}{LIT} & \fmTag{fmRIG}{RIG}\newline \fmTag{fmLIT}{LIT} & - \\
\rowcolor{altrow}
44 & \texttt{fbg-\allowbreak rail-\allowbreak axle-\allowbreak inversion} & \fmTag{fmPass}{PASS} & \fmTag{fmPass}{PASS} & \fmTag{fmPass}{PASS} & All three trials passed. \\
45 & \texttt{fbg-\allowbreak strain-\allowbreak amplifier-\allowbreak design} & \fmTag{fmINS}{INS}\newline \fmTag{fmMIS}{MIS} & \fmTag{fmINS}{INS}\newline \fmTag{fmMIS}{MIS} & \fmTag{fmINS}{INS}\newline \fmTag{fmMIS}{MIS} & Agents may have misinterpreted one task requirement, leading them to violate another. \\
\rowcolor{altrow}
46 & \texttt{heat-\allowbreak assisted-\allowbreak detection-\allowbreak and-\allowbreak ranging} & \fmTag{fmREAL}{REAL}\newline \fmTag{fmINT}{INT}\newline \fmTag{fmRIG}{RIG}\newline \fmTag{fmEXE}{EXE} & \fmTag{fmREAL}{REAL}\newline \fmTag{fmINT}{INT}\newline \fmTag{fmRIG}{RIG}\newline \fmTag{fmEXE}{EXE} & \fmTag{fmREAL}{REAL}\newline \fmTag{fmINT}{INT}\newline \fmTag{fmRIG}{RIG}\newline \fmTag{fmEXE}{EXE} & Agents overlooked emissivity structure, real sensing conditions, and range identifiability, then failed to implement stable decomposition of temperature, emissivity, texture, and distance. \\
47 & \texttt{hspic-\allowbreak spectrometer-\allowbreak e2e} & \fmTag{fmINT}{INT}\newline \fmTag{fmRIG}{RIG}\newline \fmTag{fmMIS}{MIS} & \fmTag{fmINT}{INT}\newline \fmTag{fmLIT}{LIT}\newline \fmTag{fmRIG}{RIG} & \fmTag{fmPass}{PASS}\newline \fmTag{fmRIG}{RIG} & Agents compute the spectral metric C and the residual correlation correctly, but in Trials 1 and 2 they never explain why the residual check is needed: a periodic (comb-like) response can have a narrow first crossing and still show large later correlation peaks (INT). The same two trials fail complexity\_scaling. They discuss round-trip delays and reflection-free cascades but never derive that a transmissive cascade's path order is limited by its stage count and that reflections remove this limit (INT). Trial 2 also fails positioning, because its comparison table has no filter-array entry and does not treat reconstructive spectrometers as a separate family (LIT). Trial 3 makes the periodic-response point explicit with a long-delay MZI (C = 6.630 nm\textsuperscript{-}\textsuperscript{1}, residual correlation 0.226 > 0.15) and passes. \\
\addlinespace[4pt]
\multicolumn{6}{@{}l}{\textbf{High Energy Physics}}\\*
\addlinespace[2pt]
\rowcolor{altrow}
48 & \texttt{corsika-\allowbreak air-\allowbreak shower-\allowbreak inverse} & \fmTag{fmINT}{INT}\newline \fmTag{fmLIT}{LIT} & \fmTag{fmINT}{INT}\newline \fmTag{fmLIT}{LIT} & \fmTag{fmINT}{INT}\newline \fmTag{fmLIT}{LIT} & - \\
49 & \texttt{icecube-\allowbreak pointsource-\allowbreak unbinned} & \fmTag{fmEXE}{EXE}\newline \fmTag{fmRIG}{RIG} & \fmTag{fmEXE}{EXE} & \fmTag{fmPass}{PASS}\newline \fmTag{fmINT}{INT}\newline \fmTag{fmLIT}{LIT} & All three runs build the search correctly and pass every hidden scientific re-run; the failures are at the program's edge: one delivered table inconsistent with its own fit, and a cache written into the read-only code directory in --localisation-batch mode. Trial 3 passes. \\
\rowcolor{altrow}
50 & \texttt{o3-\allowbreak monodromy-\allowbreak line-\allowbreak defect-\allowbreak mc} & \fmTag{fmPass}{PASS}\newline \fmTag{fmRIG}{RIG}\newline \fmTag{fmPHY}{PHY} & \fmTag{fmPass}{PASS}\newline \fmTag{fmRIG}{RIG}\newline \fmTag{fmPHY}{PHY} & \fmTag{fmPass}{PASS}\newline \fmTag{fmRIG}{RIG}\newline \fmTag{fmPHY}{PHY} & In the line-defect Monte Carlo study, agents obtained multiple RG fixed points but could not identify which were physically relevant. \\
51 & \texttt{o3-\allowbreak surface-\allowbreak cft-\allowbreak fuzzy-\allowbreak sphere} & \fmTag{fmRIG}{RIG}\newline \fmTag{fmPHY}{PHY} & \fmTag{fmRIG}{RIG}\newline \fmTag{fmPHY}{PHY} & \fmTag{fmRIG}{RIG}\newline \fmTag{fmPHY}{PHY} & The task hinged on distinguishing ordinary from extraordinary boundary conditions and pairing operators correctly; agents misimplemented one boundary condition from the start and misidentified the second and third S\_z=1 states. \\
\rowcolor{altrow}
52 & \texttt{sipm-\allowbreak spe-\allowbreak gain-\allowbreak breakdown} & \fmTag{fmINS}{INS}\newline \fmTag{fmLIT}{LIT} & \fmTag{fmPass}{PASS} & \fmTag{fmPass}{PASS} & - \\
\addlinespace[4pt]
\multicolumn{6}{@{}l}{\textbf{Chemical Physics}}\\*
\addlinespace[2pt]
53 & \texttt{agi-\allowbreak surface-\allowbreak hydrophobicity} & \fmTag{fmINT}{INT}\newline \fmTag{fmRIG}{RIG} & \fmTag{fmINT}{INT}\newline \fmTag{fmRIG}{RIG}\newline \fmTag{fmMIS}{MIS} & \fmTag{fmINT}{INT}\newline \fmTag{fmRIG}{RIG} & Agents avoided the intended basal interface by selecting a nonpolar prism. Cross-surface probes and system-size justification were incomplete; Trials 2--3 lacked stable estimates. Trial 3's water model conflicted with the planned study; reviewers overlooked Trial 2's conclusion. \\
\rowcolor{altrow}
54 & \texttt{hole-\allowbreak bifurcation-\allowbreak ml-\allowbreak landscape} & \fmTag{fmPass}{PASS} & \fmTag{fmRIG}{RIG}\newline \fmTag{fmLIT}{LIT} & \fmTag{fmRIG}{RIG}\newline \fmTag{fmLIT}{LIT} & Rerun after a task fix: Trial 1 passed. Trials 2 and 3 simulated many designs differently from the reference, changing high-yield labels and the train/test split; Trial 2 checked results with the model that produced them and omitted the PCCP study. \\
55 & \texttt{p2d-\allowbreak cell-\allowbreak model-\allowbreak identification} & \fmTag{fmRIG}{RIG}\newline \fmTag{fmRRMIS}{RRMIS}\newline \fmTag{fmPHY}{PHY}\newline \fmTag{fmLIT}{LIT} & \fmTag{fmRIG}{RIG}\newline \fmTag{fmINS}{INS}\newline \fmTag{fmRRMIS}{RRMIS}\newline \fmTag{fmLIT}{LIT} & \fmTag{fmRIG}{RIG}\newline \fmTag{fmRRMIS}{RRMIS}\newline \fmTag{fmPHY}{PHY}\newline \fmTag{fmLIT}{LIT} & Agents meshed the hole-covered surface too coarsely, and their models, with poorly chosen transport and kinetic prefactors, did not generalize to other currents and temperatures; the rubric grader also penalized missing measured curves that the task forbade downloading. \\
\rowcolor{altrow}
56 & \texttt{smfret-\allowbreak research-\allowbreak v5} & \fmTag{fmPHY}{PHY}\newline \fmTag{fmRIG}{RIG}\newline \fmTag{fmLIT}{LIT} & \fmTag{fmPHY}{PHY}\newline \fmTag{fmRIG}{RIG} & \fmTag{fmHAL}{HAL}\newline \fmTag{fmPHY}{PHY} & Agents used incorrect physical models and unsupported literature, and their conclusions did not match the evidence. \\
\end{longtable}
 \endgroup
 \begin{table}[H]
\centering
\caption{Regrading with a second reviewer agent. GPT-5.6 Sol (xhigh), the default reviewer agent of all main results, and Claude Opus 5.5 grade the same first-round rollouts with the same rubrics. Rubric scores are as in Figure~\ref{fig:per-task-rubric}. The pass/fail columns cover the rollouts that pass the verifier, the only ones whose outcome the reviewer agent decides; $\kappa$ is Cohen's kappa, and Opus only and Sol only count rollouts passed by just one of the two reviewer agents. Two Claude Opus 5.5 rollouts were not regraded.}
\label{tab:reviewer-regrading}
\includegraphics[width=\linewidth]{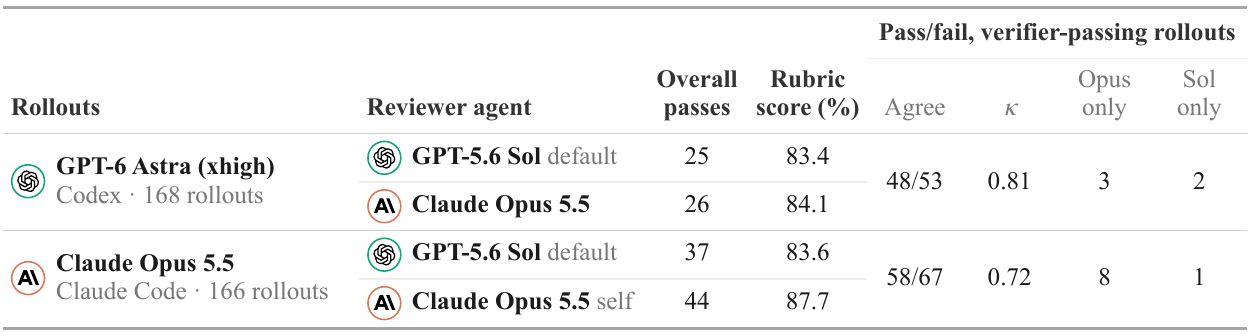}
\end{table}
 \clearpage
\section{Slot Accounting}
\label{app:result-accounting}
Table~\ref{tab:zero-filled} counts the slots without a valid result, and Table~\ref{tab:ci-main-results} gives confidence intervals for the overall pass rates from a bootstrap over tasks.

\begin{table}[htbp]
\centering
\footnotesize
\setlength{\tabcolsep}{4pt}
\caption{Slots without a valid result per configuration: no rollout, or a rollout without complete scoring. Each of them counts as a failure in the overall pass rate and as zero in the score. A rollout that ended at a time or memory limit, or that the model ended, is a scored failure and is not counted here. Every configuration has 168 slots.}
\label{tab:zero-filled}
\begin{tabular}{@{}lr@{\hspace{14pt}}lr@{\hspace{14pt}}lr@{}}
\toprule
Configuration & Slots & Configuration & Slots & Configuration & Slots \\
\midrule
Claude Opus 5.5 & \FPOpusFiveFiveZeroFilled{} & GPT-5.6 Sol (max) & \FPSolMaxZeroFilled{} & Grok 4.7 & \FPGrokFourSevenZeroFilled{} \\
Claude Fable 5.1 & \FPFableZeroFilled{} & GPT-5.6 Sol (xhigh) & \FPSolZeroFilled{} & Grok 4.6 & \FPGrokZeroFilled{} \\
Claude Opus 5 & \FPOpusFiveZeroFilled{} & GPT-5.6 Luna (max) & \FPLunaMaxZeroFilled{} & Kimi K3.1 Pro & \FPKimiThreeOneProZeroFilled{} \\
Claude Haiku 5.5 & \FPHaikuFiveFiveZeroFilled{} & GPT-5.6 Luna (xhigh) & \FPLunaZeroFilled{} & Kimi K3 & \FPKimiZeroFilled{} \\
GPT-6 Astra (max) & \FPAstraMaxZeroFilled{} & GPT-5.6 Terra (max) & \FPTerraMaxZeroFilled{} & MiMo V2.6 Pro & \FPMiMoZeroFilled{} \\
GPT-6 Astra (xhigh) & \FPAstraZeroFilled{} & GPT-5.6 Terra (xhigh) & \FPTerraZeroFilled{} & GLM-5.3 & \FPGLMZeroFilled{} \\
GPT-6.1 Sol & \FPSixOneSolZeroFilled{} & Gemini 4 Argon & \FPGeminiFourZeroFilled{} & Mistral Large 4 & \FPMistralLargeFourZeroFilled{} \\
GPT-6 Sol & \FPSixSolZeroFilled{} & Gemini 3.8 Flash & \FPGeminiZeroFilled{} & DeepSeek V4 Pro & \FPDeepSeekZeroFilled{} \\
GPT-6.1 Terra & \FPSixOneTerraZeroFilled{} & & & DeepSeek V4.1 Flash & \FPDeepSeekFourOneFlashZeroFilled{} \\
\bottomrule
\end{tabular}
\end{table}

\begin{table}[htbp]
\centering
\small
\setlength{\tabcolsep}{4pt}
\caption{Overall pass rates with task-clustered uncertainty, over all 56 tasks and 3 trials per task; unavailable outcomes count as failures, as in Table~\ref{tab:main-results}. 95\% confidence intervals from a bootstrap over tasks, 10,000 resamples, which accounts for the correlation of the three trials of a task. The Binomial column repeats the previous approximate 95\% binomial (Wald) half-width over 168 rollouts, which treats the rollouts as independent. Pass@3 is the share of tasks with at least one passing trial. 3/3 and $\geq$1/3 count the tasks (of 56) passed in all three trials and in at least one. $^\dagger$Not statistically separable from Claude Opus 5.5 (two-sided paired sign-flip permutation test over tasks, unadjusted $p\geq 0.05$).}
\label{tab:ci-main-results}
\begin{tabular}{@{}lrcrrcrr@{}}
\toprule
 & \multicolumn{3}{c}{Pass@1 (\%)} & \multicolumn{2}{c}{Pass@3 (\%)} & \multicolumn{2}{c}{Tasks passed} \\
\cmidrule(lr){2-4}\cmidrule(lr){5-6}\cmidrule(l){7-8}
Configuration & Est. & 95\% CI & Binomial & Est. & 95\% CI & 3/3 & $\geq$1/3 \\
\midrule
Claude Opus 5.5 & 27.4 & [17.9, 36.9] & $\pm$6.7 & 44.6 & [32.1, 57.1] & 8 & 25 \\
GPT-6 Astra (max)$^\dagger$ & 23.2 & [13.7, 33.9] & $\pm$6.4 & 30.4 & [17.9, 42.9] & 9 & 17 \\
GPT-6 Astra (xhigh)$^\dagger$ & 19.6 & [11.3, 28.6] & $\pm$6.0 & 30.4 & [17.9, 42.9] & 4 & 17 \\
GPT-6.1 Sol$^\dagger$ & 19.6 & [11.3, 29.2] & $\pm$6.0 & 30.4 & [19.6, 42.9] & 6 & 17 \\
GPT-6.1 Terra & 14.3 & [7.1, 22.6] & $\pm$5.3 & 21.4 & [10.7, 32.1] & 4 & 12 \\
GPT-6 Sol & 14.3 & [7.7, 21.4] & $\pm$5.3 & 28.6 & [17.9, 41.1] & 2 & 16 \\
Claude Fable 5.1 & 11.3 & [5.4, 18.5] & $\pm$4.8 & 21.4 & [10.7, 32.1] & 3 & 12 \\
Claude Opus 5 & 9.5 & [4.2, 16.1] & $\pm$4.4 & 17.9 & [8.9, 28.6] & 1 & 10 \\
Gemini 4 Argon & 8.3 & [3.6, 14.3] & $\pm$4.2 & 17.9 & [8.9, 28.6] & 1 & 10 \\
GPT-5.6 Sol (max) & 7.1 & [2.4, 13.7] & $\pm$3.9 & 12.5 & [5.4, 21.4] & 2 & 7 \\
Kimi K3.1 Pro & 6.5 & [2.4, 12.5] & $\pm$3.7 & 12.5 & [5.4, 21.4] & 1 & 7 \\
Grok 4.7 & 4.2 & [1.2, 7.7] & $\pm$3.0 & 10.7 & [3.6, 19.6] & 0 & 6 \\
Claude Haiku 5.5 & 4.2 & [1.2, 7.7] & $\pm$3.0 & 10.7 & [3.6, 19.6] & 0 & 6 \\
GPT-5.6 Sol (xhigh) & 4.2 & [1.2, 8.3] & $\pm$3.0 & 8.9 & [1.8, 17.9] & 0 & 5 \\
GLM-5.3 & 4.2 & [1.2, 8.3] & $\pm$3.0 & 8.9 & [1.8, 17.9] & 0 & 5 \\
GPT-5.6 Terra (max) & 3.6 & [0.6, 8.3] & $\pm$2.8 & 7.1 & [1.8, 14.3] & 1 & 4 \\
GPT-5.6 Luna (max) & 3.6 & [0.0, 8.9] & $\pm$2.8 & 3.6 & [0.0, 8.9] & 2 & 2 \\
DeepSeek V4.1 Flash & 3.6 & [0.0, 8.3] & $\pm$2.8 & 5.4 & [0.0, 12.5] & 1 & 3 \\
Kimi K3 & 2.4 & [0.0, 5.4] & $\pm$2.3 & 5.4 & [0.0, 12.5] & 0 & 3 \\
Grok 4.6 & 2.4 & [0.0, 6.5] & $\pm$2.3 & 3.6 & [0.0, 8.9] & 1 & 2 \\
GPT-5.6 Luna (xhigh) & 1.8 & [0.0, 4.8] & $\pm$2.0 & 3.6 & [0.0, 8.9] & 0 & 2 \\
MiMo V2.6 Pro & 1.2 & [0.0, 3.0] & $\pm$1.6 & 3.6 & [0.0, 8.9] & 0 & 2 \\
GPT-5.6 Terra (xhigh) & 1.2 & [0.0, 3.6] & $\pm$1.6 & 1.8 & [0.0, 5.4] & 0 & 1 \\
DeepSeek V4 Pro & 0.6 & [0.0, 1.8] & $\pm$1.2 & 1.8 & [0.0, 5.4] & 0 & 1 \\
Gemini 3.8 Flash & 0.0 & [0.0, 0.0] & $\pm$0.0 & 0.0 & [0.0, 0.0] & 0 & 0 \\
Mistral Large 4 & 0.0 & [0.0, 0.0] & $\pm$0.0 & 0.0 & [0.0, 0.0] & 0 & 0 \\
\bottomrule
\end{tabular}
\end{table}
  \clearpage
\section{Agent Resource Use}
\label{sec:agent-resources}
\begin{figure}[H]
\centering
\includegraphics[width=\linewidth]{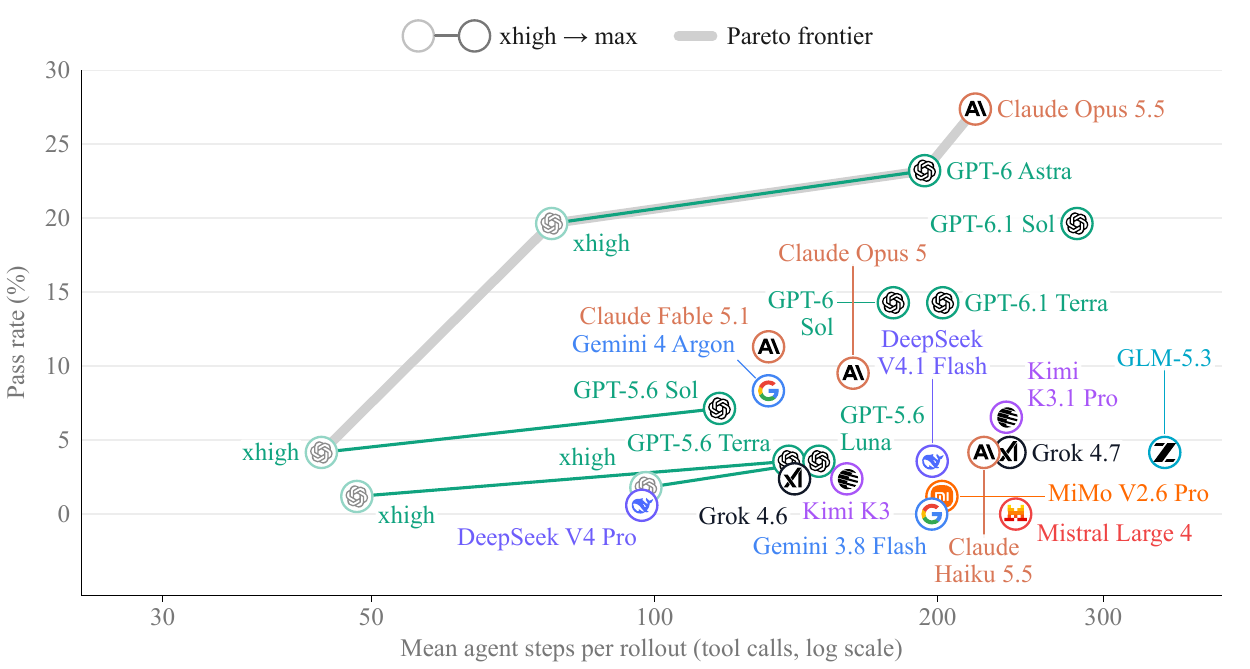}
\caption{Mean agent steps per rollout versus overall pass rate (\%). Steps count recorded tool invocations, including failed and interrupted calls.}
\label{fig:agent-steps}
\end{figure}

\begin{figure}[H]
\centering
\includegraphics[width=\linewidth]{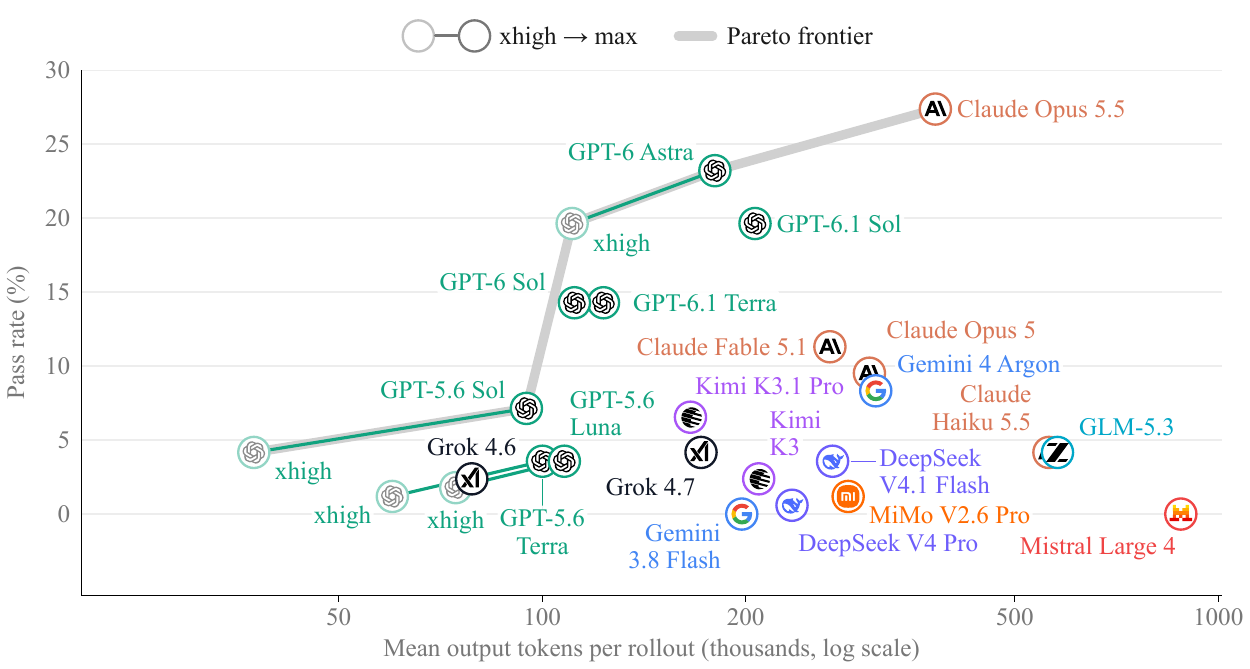}
\caption{Mean reported output tokens per rollout, in thousands, versus overall pass rate (\%). Token counts follow provider or harness usage accounting. }
\label{fig:output-tokens}
\end{figure}

\begin{figure}[p]
\centering
\includegraphics[width=\linewidth]{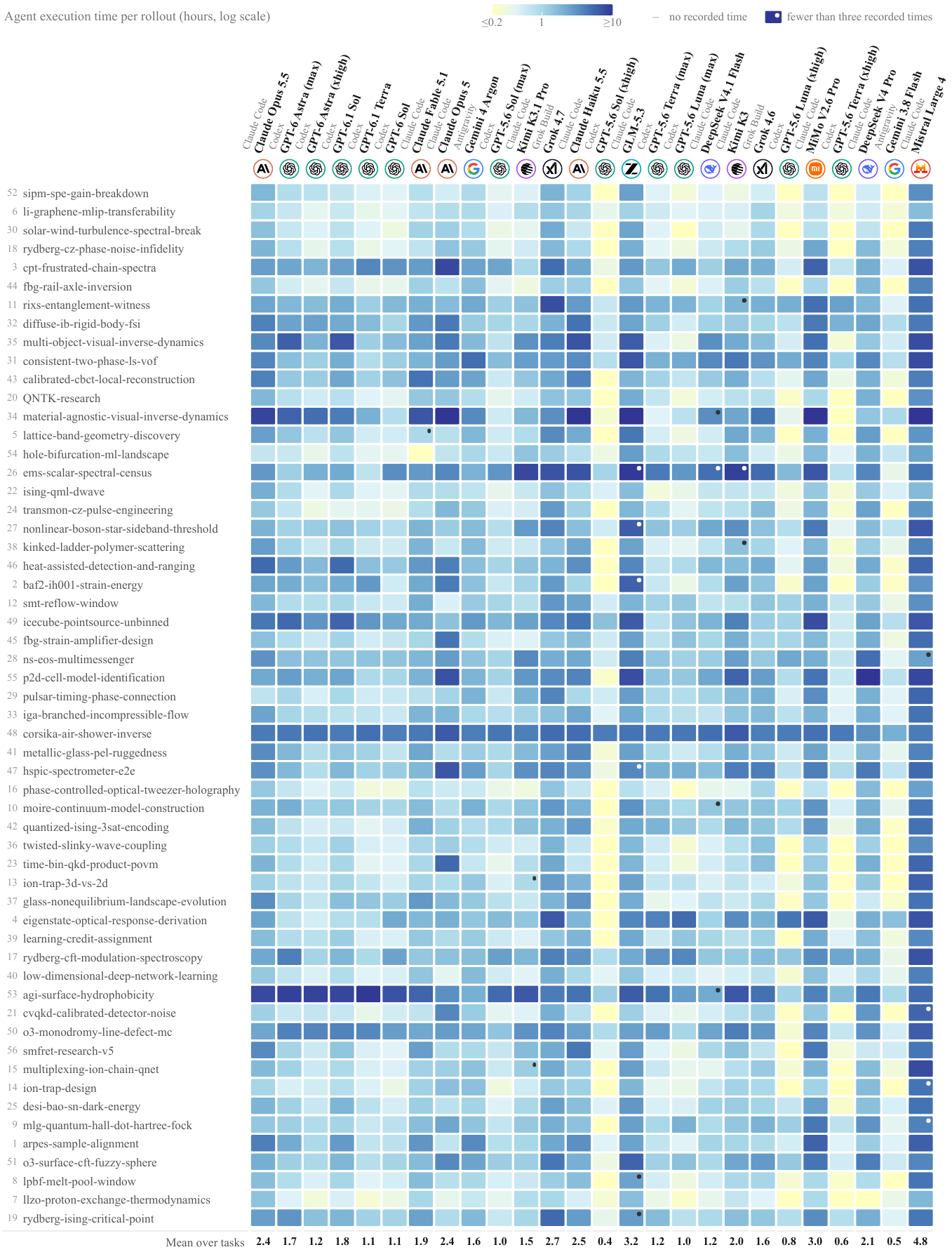}
\caption{Agent execution time per rollout, in hours, per task and model: the mean over the recorded rollouts in each cell. The bottom row gives each model's mean over tasks with records, as plotted in Figure~\ref{fig:agent-time}. A dash marks a task without a recorded time and a dot one with fewer than three.}
\label{fig:per-task-agent-time}
\end{figure}

\begin{figure}[p]
\centering
\includegraphics[width=\linewidth]{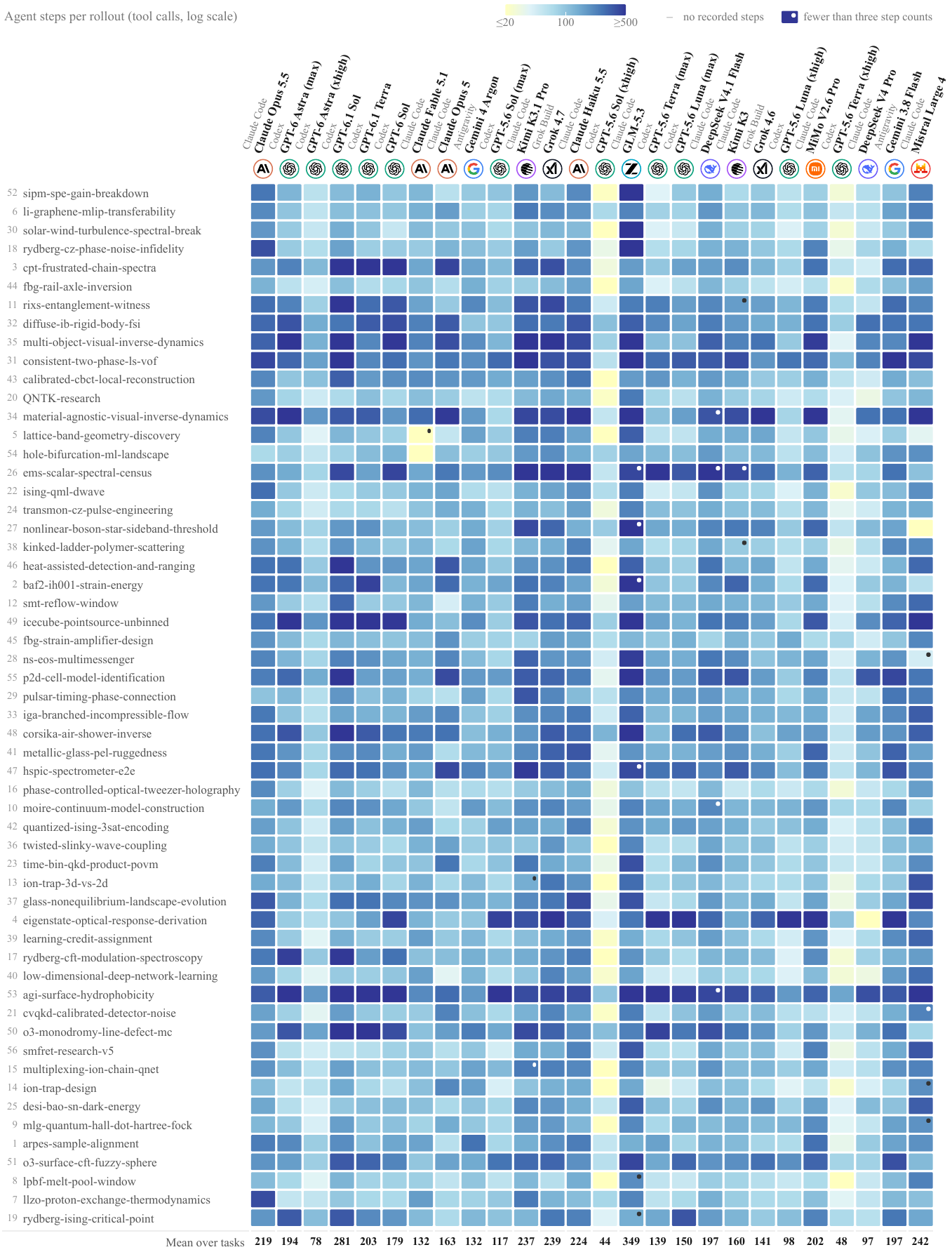}
\caption{Agent steps (tool calls) per rollout, per task and model: the mean over the recorded rollouts in each cell. The bottom row gives each model's mean over tasks with records, as plotted in Figure~\ref{fig:agent-steps}. Rows follow Figure~\ref{fig:per-task-verifier}. A dash marks a task without a recorded step count and a dot one with fewer than three.}
\label{fig:per-task-agent-steps}
\end{figure}

\begin{figure}[p]
\centering
\includegraphics[width=\linewidth]{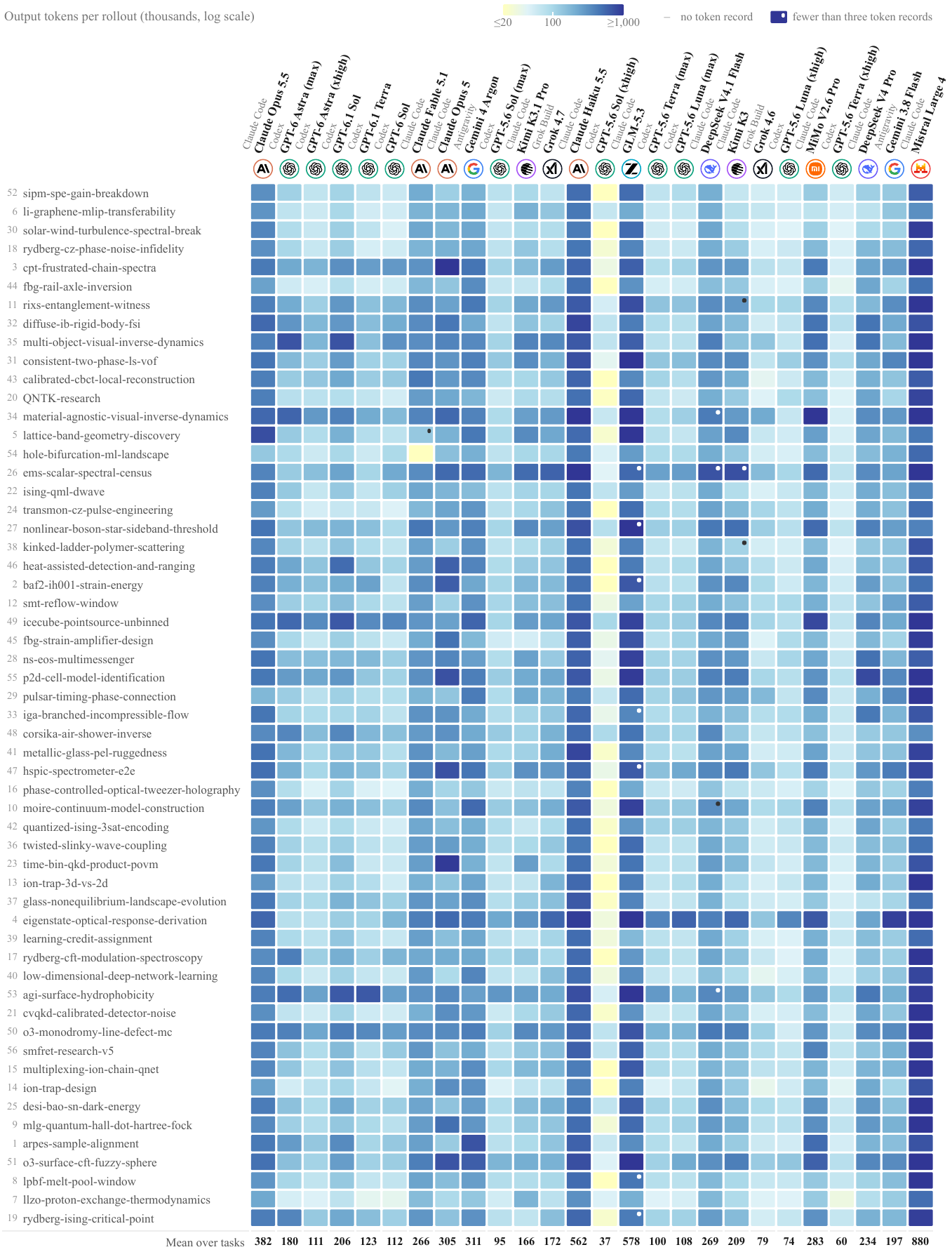}
\caption{Output tokens per rollout, in thousands, per task and model: the mean over the recorded rollouts in each cell. The bottom row gives each model's mean over tasks with records, as plotted in Figure~\ref{fig:output-tokens}. Rows follow Figure~\ref{fig:per-task-verifier}. A dash marks a task without an output-token record and a dot one with fewer than three.}
\label{fig:per-task-output-tokens}
\end{figure}

Figures~\ref{fig:agent-time}, \ref{fig:agent-steps}, and~\ref{fig:output-tokens} compare overall pass rate (Table~\ref{tab:main-results}, overall Pass@1) with agent execution time, agent steps, and output tokens per rollout. Each resource is averaged within each task and then equally across tasks with recorded values, using the same selected rollouts as Table~\ref{tab:main-results}. A step denotes one tool invocation. The gray bands mark the empirical Pareto frontiers for lower resource use and higher pass rate. Figures~\ref{fig:per-task-agent-time}--\ref{fig:per-task-output-tokens} show how each resource is distributed over tasks and models, in the layout of Figures~\ref{fig:per-task-verifier}--\ref{fig:per-task-score}. Their colors follow a logarithmic scale, and the few values beyond its ends take the end colors. Appendix~\ref{app:resource-accounting} gives the measurement definitions and coverage.

 \subsection{Agent Resource Accounting}

\label{app:resource-accounting}

The resource plots use the same selected rollouts and 56-task overall pass rate as the cost comparison. For each resource, available values are averaged within each task and then equally across tasks. Unavailable counters are omitted from resource means, not replaced with zero. Table~\ref{tab:resource-coverage} reports the number of rollouts and tasks contributing to each resource estimate.

Agent time is the recorded execution-stage wall-clock duration, excluding environment setup, verification and rubric review. When an idle-timeout record lacks that stage timer, its recorded elapsed execution time supplies the duration. Agent steps count recorded tool invocations, including failed and interrupted calls. A partial trace is used only when a termination monitor supplies a tool-call counter. Output tokens use the reported agent output-token count; unavailable usage records are omitted. Tokenization and output-token accounting can differ between providers. In Claude Code runs, agent time and steps include subagent work.

\begin{table}[htbp]\centering\small

\renewcommand{\arraystretch}{1.2}

\caption{Resource coverage: contributing rollouts / tasks for each model.}

\label{tab:resource-coverage}

\begin{tabular}{lrrr}\toprule

Model & Agent time & Agent steps & Output tokens \\\midrule

Claude Opus 5.5 & 168 / 56 & 168 / 56 & 168 / 56 \\

GPT-6 Astra (max) & 168 / 56 & 168 / 56 & 168 / 56 \\

GPT-6 Astra (xhigh) & 168 / 56 & 168 / 56 & 168 / 56 \\

GPT-6.1 Sol & 168 / 56 & 168 / 56 & 168 / 56 \\

GPT-6.1 Terra & 168 / 56 & 168 / 56 & 168 / 56 \\

GPT-6 Sol & 168 / 56 & 168 / 56 & 168 / 56 \\

Claude Fable 5.1 & 166 / 56 & 166 / 56 & 166 / 56 \\

Claude Opus 5 & 168 / 56 & 168 / 56 & 168 / 56 \\

Gemini 4 Argon & 168 / 56 & 168 / 56 & 168 / 56 \\

GPT-5.6 Sol (max) & 168 / 56 & 168 / 56 & 168 / 56 \\

Kimi K3.1 Pro & 165 / 56 & 165 / 56 & 168 / 56 \\

Grok 4.7 & 168 / 56 & 168 / 56 & 168 / 56 \\

Claude Haiku 5.5 & 168 / 56 & 168 / 56 & 168 / 56 \\

GPT-5.6 Sol (xhigh) & 168 / 56 & 168 / 56 & 168 / 56 \\

GLM-5.3 & 160 / 56 & 160 / 56 & 159 / 56 \\

GPT-5.6 Terra (max) & 168 / 56 & 168 / 56 & 168 / 56 \\

GPT-5.6 Luna (max) & 168 / 56 & 168 / 56 & 168 / 56 \\

DeepSeek V4.1 Flash & 162 / 56 & 162 / 56 & 162 / 56 \\

Kimi K3 & 165 / 56 & 165 / 56 & 165 / 56 \\

Grok 4.6 & 168 / 56 & 168 / 56 & 168 / 56 \\

GPT-5.6 Luna (xhigh) & 168 / 56 & 168 / 56 & 168 / 56 \\

MiMo V2.6 Pro & 168 / 56 & 168 / 56 & 168 / 56 \\

GPT-5.6 Terra (xhigh) & 168 / 56 & 168 / 56 & 168 / 56 \\

DeepSeek V4 Pro & 168 / 56 & 168 / 56 & 168 / 56 \\

Gemini 3.8 Flash & 168 / 56 & 168 / 56 & 168 / 56 \\

Mistral Large 4 & 164 / 56 & 164 / 56 & 168 / 56 \\

\bottomrule\end{tabular}\end{table}

\begin{table}[htbp]
\centering
\caption{Resource use per rollout for each model configuration: mean, median (Med.) and 95th percentile (P95), pooled over the rollouts of Table~\ref{tab:main-results} that record the value. These are the same selected rollouts as Figures~\ref{fig:pareto-frontier}, \ref{fig:agent-steps} and~\ref{fig:output-tokens}, and rows follow Table~\ref{tab:main-results}. Tool calls count recorded tool invocations, including failed and interrupted calls. A step is one model call of the agent that returned a response, and one step can make several tool calls or none. Time is the execution-stage wall-clock duration in hours, excluding environment setup, verification and rubric review; because it is wall-clock time, slow provider routes and rate limiting inflate it. Cost is the reported inference cost per rollout in US dollars, as in Figure~\ref{fig:pareto-frontier} (metered, priced from recorded tokens at list prices, or estimated), and output tokens are in millions. Unavailable values are omitted, not set to zero; $^{\dagger}$ marks a statistic over fewer than all 168 designated rollouts of the configuration (Table~\ref{tab:resource-coverage} gives the coverage of time, tool calls and output tokens). P95 and the median use linear interpolation between order statistics (NumPy's default). The figures instead average within each task and then across tasks; both means coincide wherever all 168 rollouts are recorded. In Claude Code runs, tool calls, steps and time include subagent work.}
\label{tab:resource-stats}
\small
\setlength{\tabcolsep}{4pt}
\resizebox{\ifdim\width>\linewidth\linewidth\else\width\fi}{!}{\begin{tabular}{l rrr rrr rrr rrr rr}
\toprule
 & \multicolumn{3}{c}{Tool calls} & \multicolumn{3}{c}{Steps} & \multicolumn{3}{c}{Time (h)} & \multicolumn{3}{c}{Cost (\$)} & \multicolumn{2}{c}{\begin{tabular}[b]{@{}c@{}}Output\\tokens (M)\end{tabular}} \\
\cmidrule(lr){2-4}\cmidrule(lr){5-7}\cmidrule(lr){8-10}\cmidrule(lr){11-13}\cmidrule(lr){14-15}
Model & Mean & Med. & P95 & Mean & Med. & P95 & Mean & Med. & P95 & Mean & Med. & P95 & Mean & Med. \\
\midrule
Claude Opus 5.5 & 219 & 204 & 371 & 191 & 168 & 339 & 2.44 & 2.06 & 5.38 & 26.95 & 24.86 & 51.89 & 0.38 & 0.39 \\
GPT-6 Astra (max) & 194 & 111 & 585 & 108 & 61 & 335 & 1.73 & 1.05 & 5.28 & 26.91 & 15.24 & 90.77 & 0.18 & 0.13 \\
GPT-6 Astra (xhigh) & 78 & 65 & 172 & 86 & 65 & 197 & 1.24 & 0.75 & 3.72 & 19.21 & 11.60 & 56.87 & 0.11 & 0.09 \\
GPT-6.1 Sol & 281 & 170 & 882 & 158 & 90 & 583 & 1.79 & 1.17 & 5.21 & 5.33 & 3.20 & 17.51 & 0.21 & 0.15 \\
GPT-6.1 Terra & 203 & 124 & 461 & 203 & 119 & 553 & 1.14 & 0.65 & 3.90 & 3.46 & 1.23 & 14.25 & 0.12 & 0.09 \\
GPT-6 Sol & 179 & 114 & 501 & 194 & 110 & 634 & 1.08 & 0.63 & 3.57 & 6.38 & 3.22 & 22.09 & 0.11 & 0.08 \\
Claude Fable 5.1 & 134\rlap{$^{\dagger}$} & 128\rlap{$^{\dagger}$} & 230\rlap{$^{\dagger}$} & 81 & 66 & 190 & 1.87\rlap{$^{\dagger}$} & 1.58\rlap{$^{\dagger}$} & 4.40\rlap{$^{\dagger}$} & 32.63 & 26.42 & 48.65 & 0.27\rlap{$^{\dagger}$} & 0.27\rlap{$^{\dagger}$} \\
Claude Opus 5 & 163 & 116 & 425 & 156 & 105 & 422 & 2.40 & 1.35 & 7.58 & 56.77 & 32.85 & 163.37 & 0.30 & 0.22 \\
Gemini 4 Argon & 132 & 122 & 225 & 123 & 119 & 207 & 1.61 & 1.41 & 3.20 & 8.87 & 7.10 & 17.39 & 0.31 & 0.28 \\
GPT-5.6 Sol (max) & 117 & 89 & 260 & 154 & 106 & 401 & 0.97 & 0.62 & 2.77 & 10.35 & 6.93 & 29.21 & 0.09 & 0.09 \\
Kimi K3.1 Pro & 238\rlap{$^{\dagger}$} & 195\rlap{$^{\dagger}$} & 530\rlap{$^{\dagger}$} & 187 & 140 & 450 & 1.52\rlap{$^{\dagger}$} & 1.01\rlap{$^{\dagger}$} & 4.08\rlap{$^{\dagger}$} & 9.60 & 7.03 & 24.43 & 0.17 & 0.13 \\
Grok 4.7 & 239 & 212 & 523 & 169 & 140 & 374 & 2.75 & 2.32 & 6.40 & 12.19 & 10.13 & 27.34 & 0.17 & 0.14 \\
Claude Haiku 5.5 & 224 & 187 & 476 & 214 & 176 & 451 & 2.50 & 1.79 & 6.39 & 6.66\rlap{$^{\dagger}$} & 5.13\rlap{$^{\dagger}$} & 15.27\rlap{$^{\dagger}$} & 0.56 & 0.52 \\
GPT-5.6 Sol (xhigh) & 44 & 34 & 102 & 66 & 55 & 168 & 0.42 & 0.27 & 1.33 & 2.79 & 1.77 & 8.43 & 0.04 & 0.03 \\
GLM-5.3 & 347\rlap{$^{\dagger}$} & 261\rlap{$^{\dagger}$} & 1,015\rlap{$^{\dagger}$} & 324 & 242 & 994 & 3.14\rlap{$^{\dagger}$} & 2.46\rlap{$^{\dagger}$} & 7.02\rlap{$^{\dagger}$} & 12.81 & 10.05 & 33.40 & 0.57\rlap{$^{\dagger}$} & 0.47\rlap{$^{\dagger}$} \\
GPT-5.6 Terra (max) & 139 & 98 & 419 & 165 & 110 & 501 & 1.15 & 0.73 & 3.89 & 5.80 & 3.73 & 16.81 & 0.10 & 0.09 \\
GPT-5.6 Luna (max) & 150 & 103 & 401 & 197 & 124 & 525 & 1.04 & 0.57 & 3.44 & 0.68 & 0.42 & 1.78 & 0.11 & 0.09 \\
DeepSeek V4.1 Flash & 191\rlap{$^{\dagger}$} & 166\rlap{$^{\dagger}$} & 404\rlap{$^{\dagger}$} & 178\rlap{$^{\dagger}$} & 159\rlap{$^{\dagger}$} & 394\rlap{$^{\dagger}$} & 1.17\rlap{$^{\dagger}$} & 0.83\rlap{$^{\dagger}$} & 3.13\rlap{$^{\dagger}$} & 0.65\rlap{$^{\dagger}$} & 0.54\rlap{$^{\dagger}$} & 1.34\rlap{$^{\dagger}$} & 0.26\rlap{$^{\dagger}$} & 0.24\rlap{$^{\dagger}$} \\
Kimi K3 & 159\rlap{$^{\dagger}$} & 123\rlap{$^{\dagger}$} & 358\rlap{$^{\dagger}$} & 153 & 122 & 382 & 1.91\rlap{$^{\dagger}$} & 1.24\rlap{$^{\dagger}$} & 5.51\rlap{$^{\dagger}$} & 9.33 & 6.64 & 22.04 & 0.21\rlap{$^{\dagger}$} & 0.16\rlap{$^{\dagger}$} \\
Grok 4.6 & 141 & 122 & 315 & 83 & 71 & 179 & 1.63 & 1.27 & 4.02 & 9.21\rlap{$^{\dagger}$} & 7.35\rlap{$^{\dagger}$} & 19.75\rlap{$^{\dagger}$} & 0.08 & 0.06 \\
GPT-5.6 Luna (xhigh) & 98 & 72 & 224 & 141 & 90 & 414 & 0.75 & 0.39 & 2.18 & 0.41 & 0.24 & 1.10 & 0.07 & 0.06 \\
MiMo V2.6 Pro & 202 & 148 & 497 & 182 & 133 & 434 & 3.01 & 2.13 & 7.47 & 0.66 & 0.47 & 1.56 & 0.28 & 0.20 \\
GPT-5.6 Terra (xhigh) & 48 & 42 & 93 & 70 & 52 & 168 & 0.58 & 0.31 & 2.02 & 2.22 & 1.53 & 5.92 & 0.06 & 0.06 \\
DeepSeek V4 Pro & 97 & 66 & 237 & 90 & 63 & 214 & 2.10 & 2.00 & 5.01 & 7.82 & 5.86 & 20.95 & 0.23 & 0.19 \\
Gemini 3.8 Flash & 197 & 157 & 433 & 207 & 165 & 447 & 0.45 & 0.36 & 1.06 & 3.69\rlap{$^{\dagger}$} & 2.73\rlap{$^{\dagger}$} & 8.38\rlap{$^{\dagger}$} & 0.20 & 0.16 \\
Mistral Large 4 & 244\rlap{$^{\dagger}$} & 205\rlap{$^{\dagger}$} & 634\rlap{$^{\dagger}$} & 205 & 173 & 562 & 4.82\rlap{$^{\dagger}$} & 4.02\rlap{$^{\dagger}$} & 9.25\rlap{$^{\dagger}$} & 10.67 & 9.29 & 24.39 & 0.88 & 0.69 \\
\bottomrule
\end{tabular}}
\end{table}
  
\clearpage
\section{Reference-material Evaluation}
\label{app:reference-material-evaluation}

\begingroup
\footnotesize
\setlength{\tabcolsep}{3pt}
\setlength{\LTleft}{0pt}
\setlength{\LTright}{0pt}
\setlength{\LTcapwidth}{\textwidth}
\renewcommand{\arraystretch}{1.15}
\definecolor{refDeduction}{HTML}{9A6B2C}
\definecolor{refFail}{HTML}{B14545}

\begin{longtable}{@{}
  >{\raggedright\arraybackslash}p{0.70\textwidth}
  >{\centering\arraybackslash}p{0.16\textwidth}
  >{\centering\arraybackslash}p{\dimexpr0.14\textwidth-4\tabcolsep\relax}
@{}}

\caption{Evaluating rubric scores for all 33 tasks where the authors provided a manuscript or paper as a reference. Only the rubrics, that actually evaluated/ needed the present manuscript were evaluated. If no blocking rubric applied to the manuscript, it is shown as a dash. If no scoring rubric applied to the manuscript it is shown as NA.}
\label{tab:reference-material-evaluation}\\

\toprule
\textbf{Task / reference} & \textbf{Weighted score} & \textbf{Blockers} \\
\midrule
\endfirsthead

\caption[]{Reference-material rubric scores (continued).}\\
\toprule
\textbf{Task / reference} & \textbf{Weighted score} & \textbf{Blockers} \\
\midrule
\endhead

\midrule
\multicolumn{3}{r}{Continued on next page}\\
\endfoot

\bottomrule
\endlastfoot

QNTK-\allowbreak research & 100\% & 2/2 \\
\rowcolor{altrow}
consistent-\allowbreak two-\allowbreak phase-\allowbreak ls-\allowbreak vof & 100\% & 1/1 \\
cpt-\allowbreak frustrated-\allowbreak chain-\allowbreak spectra & N/A & \textcolor{refFail}{\textbf{0/1 - fail}} \\
\rowcolor{altrow}
cvqkd-\allowbreak calibrated-\allowbreak detector-\allowbreak noise & 100\% & \textcolor{refFail}{\textbf{2/3 - fail}} \\
diffuse-\allowbreak ib-\allowbreak rigid-\allowbreak body-\allowbreak fsi & 100\% & - \\
\rowcolor{altrow}
eigenstate-\allowbreak optical-\allowbreak response-\allowbreak derivation & 100\% & 5/5 \\
fbg-\allowbreak rail-\allowbreak axle-\allowbreak inversion & 100\% & - \\
\rowcolor{altrow}
fbg-\allowbreak strain-\allowbreak amplifier-\allowbreak design & 100\% & 2/2 \\
glass-\allowbreak nonequilibrium-\allowbreak landscape-\allowbreak evolution - bundled report & \textcolor{refDeduction}{\textbf{91.94\%}} & 2/2 \\
\rowcolor{altrow}
glass-\allowbreak nonequilibrium-\allowbreak landscape-\allowbreak evolution - original paper & \textcolor{refDeduction}{\textbf{50\%}} & - \\
heat-\allowbreak assisted-\allowbreak detection-\allowbreak and-\allowbreak ranging & 100\% & - \\
\rowcolor{altrow}
hole-\allowbreak bifurcation-\allowbreak ml-\allowbreak landscape & \textcolor{refDeduction}{\textbf{81.58\%}} & - \\
hspic-\allowbreak spectrometer-\allowbreak e2e & 100\% & 3/3 \\
\rowcolor{altrow}
icecube-\allowbreak pointsource-\allowbreak unbinned & 100\% & - \\
ion-\allowbreak trap-\allowbreak 3d-\allowbreak vs-\allowbreak 2d - bundled report & 100\% & 4/4 \\
\rowcolor{altrow}
ion-\allowbreak trap-\allowbreak 3d-\allowbreak vs-\allowbreak 2d - original paper & 100\% & 1/1 \\
ion-\allowbreak trap-\allowbreak design - bundled report & 100\% & - \\
\rowcolor{altrow}
ion-\allowbreak trap-\allowbreak design - original paper & 100\% & - \\
lattice-\allowbreak band-\allowbreak geometry-\allowbreak discovery & 100\% & 1/1 \\
\rowcolor{altrow}
learning-\allowbreak credit-\allowbreak assignment & 100\% & 4/4 \\
li-\allowbreak graphene-\allowbreak mlip-\allowbreak transferability & \textcolor{refDeduction}{\textbf{93.52\%}} & 2/2 \\
\rowcolor{altrow}
llzo-\allowbreak proton-\allowbreak exchange-\allowbreak thermodynamics & 100\% & - \\
low-\allowbreak dimensional-\allowbreak deep-\allowbreak network-\allowbreak learning & 100\% & 7/7 \\
\rowcolor{altrow}
material-\allowbreak agnostic-\allowbreak visual-\allowbreak inverse-\allowbreak dynamics & 100\% & - \\
metallic-\allowbreak glass-\allowbreak pel-\allowbreak ruggedness & \textcolor{refDeduction}{\textbf{72.73\%}} & 1/1 \\
\rowcolor{altrow}
moire-\allowbreak continuum-\allowbreak model-\allowbreak construction & \textcolor{refDeduction}{\textbf{80\%}} & - \\
multi-\allowbreak object-\allowbreak visual-\allowbreak inverse-\allowbreak dynamics & 100\% & - \\
\rowcolor{altrow}
multiplexing-\allowbreak ion-\allowbreak chain-\allowbreak qnet & 100\% & - \\
o3-\allowbreak surface-\allowbreak cft-\allowbreak fuzzy-\allowbreak sphere & 100\% & - \\
\rowcolor{altrow}
phase-\allowbreak controlled-\allowbreak optical-\allowbreak tweezer-\allowbreak holography & N/A & \textcolor{refFail}{\textbf{0/1 - fail}} \\
rixs-\allowbreak entanglement-\allowbreak witness & 100\% & - \\
\rowcolor{altrow}
rydberg-\allowbreak cft-\allowbreak modulation-\allowbreak spectroscopy & 100\% & 4/4 \\
rydberg-\allowbreak cz-\allowbreak phase-\allowbreak noise-\allowbreak infidelity & 100\% & 6/6 \\
\rowcolor{altrow}
rydberg-\allowbreak ising-\allowbreak critical-\allowbreak point & 100\% & 3/3 \\
sipm-\allowbreak spe-\allowbreak gain-\allowbreak breakdown & 100\% & - \\
\rowcolor{altrow}
time-\allowbreak bin-\allowbreak qkd-\allowbreak product-\allowbreak povm & 100\% & 5/5 \\

\end{longtable}
\endgroup

\end{document}